\documentclass[lettersize,journal]{IEEEtran}
\usepackage{amsmath,amsfonts}
\usepackage{algorithmic}
\usepackage{algorithm}
\usepackage{array}
\usepackage[caption=false,font=normalsize,labelfont=sf,textfont=sf]{subfig}
\usepackage{textcomp}
\usepackage{stfloats}
\usepackage{url}
\usepackage{verbatim}
\usepackage{graphicx}
\usepackage{cite}
\usepackage{makecell}
\usepackage{utfsym}
\usepackage{multirow}
\usepackage{colortbl}
\usepackage{amssymb}
\usepackage{url}
\usepackage{booktabs}

\begin{document}

\title{Image-Text-Image Knowledge Transfer \\ for Lifelong Person Re-Identification \\ with Hybrid Clothing States}

\author{
Qizao Wang, Xuelin Qian, Bin Li, Yanwei Fu, and Xiangyang Xue 
\thanks{Qizao Wang is with the School of Automation, Northwestern Polytechnical University, Xi'an 710021, China, and also with the School of Computer Science, Shanghai Key Lab of Intelligent Information Processing, Fudan University, Shanghai 200437, China. Email: qzwang22@m.fudan.edu.cn.}
\thanks{Xuelin Qian is with the School of Automation, Northwestern Polytechnical University, Xi'an 710021, China, and also with the Shenzhen Research Institute of Northwestern Polytechnical University, Shenzhen 518057, China. Email: xlqian@nwpu.edu.cn. 
Xuelin Qian is the corresponding author.
}
\thanks{Bin Li and Xiangyang Xue are with the School of Computer Science, and Shanghai Key Lab of Intelligent Information Processing, Fudan University, Shanghai 200437, China. Email: \{libin, xyxue\}@fudan.edu.cn.}
\thanks{Yanwei Fu is with the School of Data Science, and Fudan ISTBI—ZJNU Algorithm Centre for Brain-inspired Intelligence, Fudan University, Shanghai 200437, China. Email: yanweifu@fudan.edu.cn.}
}

\maketitle

\begin{abstract}
With the continuous expansion of intelligent surveillance networks, lifelong person re-identification (LReID) has received widespread attention, pursuing the need of self-evolution across different domains. However, existing LReID studies accumulate knowledge with the assumption that people would not change their clothes. In this paper, we propose a more practical task, namely lifelong person re-identification with hybrid clothing states (LReID-Hybrid), which takes a series of cloth-changing and same-cloth domains into account during lifelong learning. To tackle the challenges of knowledge granularity mismatch and knowledge presentation mismatch in LReID-Hybrid, we take advantage of the consistency and generalization capabilities of the text space, and propose a novel framework, dubbed \textit{Teata}, to effectively align, transfer, and accumulate knowledge in an ``image-text-image'' closed loop. Concretely, to achieve effective knowledge transfer, we design a Structured Semantic Prompt (SSP) learning to decompose the text prompt into several structured pairs to distill knowledge from the image space with a unified granularity of text description. Then, we introduce a Knowledge Adaptation and Projection (KAP) strategy, which tunes text knowledge via a slow-paced learner to adapt to different tasks without catastrophic forgetting. Extensive experiments demonstrate the superiority of our proposed \textit{Teata} for LReID-Hybrid as well as on conventional LReID benchmarks over advanced methods.
\end{abstract}

\begin{IEEEkeywords}
Person Re-Identification, Lifelong Learning, Hybrid Clothing States.
\end{IEEEkeywords}

\begin{figure*}[t]
   \centering
    \includegraphics[width=1\linewidth]{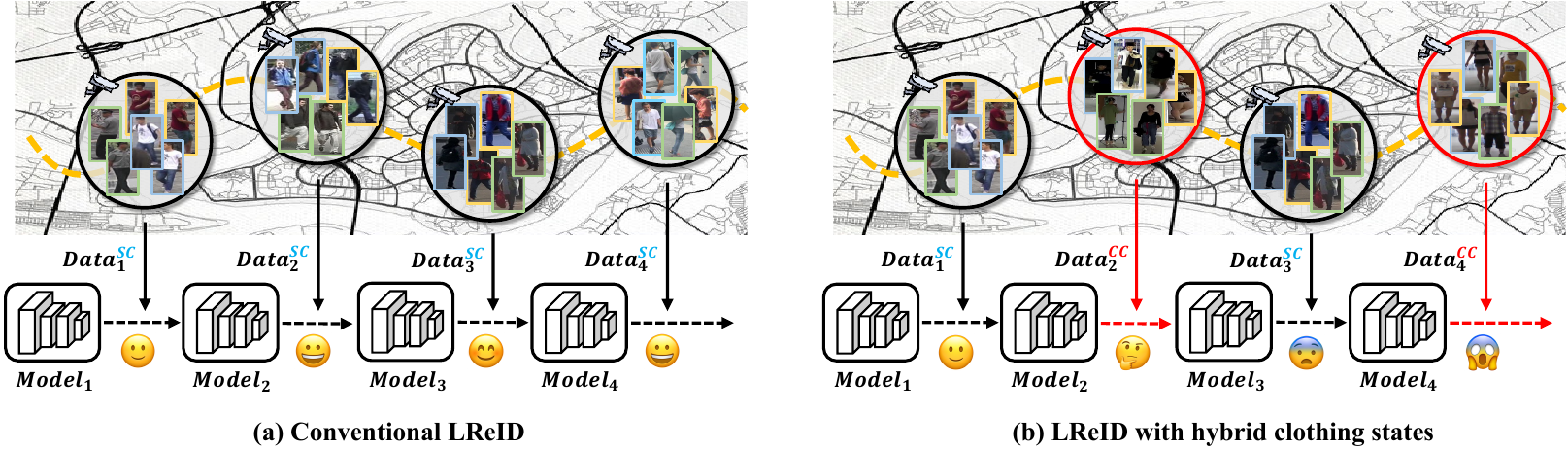}
    \caption{\textbf{Comparison between the conventional LReID task and our proposed LReID with hybrid clothing states (LReID-Hybrid).} 
    Person images from the same dataset with the same bounding box colour share the same identity. ``\textcolor[RGB]{79,193,234}{SC}'' and ``\textcolor{red}{CC}'' denote domains with same-cloth and cloth-changing states, respectively. Emojis represent the attitudes of the model towards acquiring new knowledge.
    Conventional lifelong ReID task (LReID) assumes there are no clothing changes for each person throughout the lifelong evolution process. This allows shared appearance knowledge to be acquired for ReID across different domains, and thus the process of acquiring new knowledge is joyful.
    We propose a more practical task that takes the clothing changes into account. The mismatch of knowledge caused by varying clothing states across domains can exacerbate knowledge forgetting, making the process of acquiring new knowledge confusing.
    }
    \label{fig:intro}
\end{figure*}

\section{Introduction}
\label{sec:intro}
\IEEEPARstart{L}{ifelong} Person Re-identification (LReID) is a sub-field of person ReID, which aims to retrieve the same individual that appeared in different camera views, while incrementally accumulating the ReID knowledge in different scenarios \cite{pu2021lifelong}. 
This topic offers numerous meaningful application scenarios, capable of accommodating the ongoing expansion of surveillance networks and facilitating the continual learning of ReID models across diverse domains. Consequently, it has increasingly captivated attention and research efforts. 
For example, Ge \textit{et al.}~\cite{ge2022lifelong} replay old samples saved in memory with specific-designed supervisions to balance stability and plasticity for lifelong learning. Yu \textit{et al.}~\cite{yu2023lifelong} further introduce knowledge refreshing and consolidation techniques to achieve positive forward and backward transfer. To avoid data privacy, Lu \textit{et al.}~\cite{lu2022augmented} present dream memory to preserve knowledge, while learning incrementally with augmented geometric distillation.

Under the efforts of existing work, shared appearance knowledge for ReID can be accumulated across different domains in the conventional LReID setting, as shown in Fig.~\ref{fig:intro}~(a). Unfortunately, they overlook a crucial factor that is common and of concern in the ReID application scenario, \textit{i.e.}, clothing changes \cite{gu2022clothes,qian2020long,jin2022cloth,yang2019person}. They follow the same-cloth ReID setting, where each person always wears the same clothes, so the acquired knowledge is mostly related to clothing appearances. When pedestrians change their clothes, the ReID models will struggle to achieve satisfactory results~\cite{wang2022co,gu2022clothes,wang2024exploring}. Similarly, in the lifelong learning scenario, the assumption about the clothing state is not conducive to the generalization and robustness of the model. For example, models could be easily deceived by people with clothing changes (CC) even after sufficient steps of continual learning on conventional same-cloth data, since the acquired knowledge has the bias that the same identity keeps the same clothes (SC).

Therefore, we advocate taking the clothing changes into account in LReID. Considering that other coexisting factors, such as camera view changes and background changes, are still thorny and cannot be ignored, we formulate a challenging LReID setting using datasets with different clothing states based on existing LReID research~\cite{pu2021lifelong,ge2022lifelong,yu2023lifelong}. Specifically, we propose a more practical LReID task, namely \textit{lifelong person re-identification with hybrid clothing states} (LReID-Hybrid), which requires constantly updating the model on a series of cloth-changing and conventional same-cloth ReID datasets, as depicted in Fig.~\ref{fig:intro}~(b). In the LReID-Hybrid setting, the model is expected to resolve the conflict of knowledge acquired when the clothing states of data alternate, and accumulate shareable and complementary knowledge from hybrid data that comes separately and incrementally. Eventually, the learned model is required to be generalized across various scenarios with different clothing conditions.

However, developing a lifelong ReID model on sequential domains with clothing state differences is a non-trivial task and inevitably encounters several challenges. 
The first one is \textit{knowledge representation mismatch}. Although the incremental datasets all belong to the task of person ReID, they naturally exhibit distribution differences due to variations in location and time, causing visual representation shifts between different learning stages. This mismatch is exacerbated by changes in the state of people's clothing. Clothing features, which serve as discriminative representations in conventional ReID datasets, become disruptive when the same person changes clothes (\textit{i.e.}, cloth-changing scenarios). Such a representation mismatch further exacerbates catastrophic forgetting of knowledge, leading to fluctuations in the performance of the model during lifelong learning.
The second challenge is \textit{knowledge granularity mismatch}. In contrast to conventional ReID, cloth-changing ReID typically requires models to perceive more fine-grained cues in person images~\cite{wang2024exploring}, since visual appearance information, which takes up a greater proportion, can be conversely misleading. The difference in representation granularity is data-driven and has no impact on learning in a single data domain.  Nevertheless, it occurs when carrying lifelong learning across hybrid domains, which inevitably hinders effective knowledge transfer and accumulation.
The above two challenges can be especially tricky when data with hybrid clothing states comes separately and sequentially in the lifelong learning setting.

In light of the great success of visual language models in effective representation~\cite{radford2021learning}, as well as their advanced applications in ReID \cite{li2023clip}, we thus seek a more compact and robust representation of person images in the text space. Natural language can intuitively and easily describe various features of person images, including appearance, gender, behavior, and more, some of which are conducive to re-identifying individuals. More importantly, natural language employs a unified sentence pattern or vocabulary when describing person images, effectively bypassing the domain differences inherent in image pixels. Based on the above inspiration, we present a novel framework, dubbed \textit{Teata}, that takes full advantage of \textbf{Te}xt modality to \textbf{a}lign, \textbf{t}ransfer and \textbf{a}ccumulate knowledge. Our core idea is a two-stage process of alternating iterations, to first distill pixel-level information and category labels into text semantic embeddings, and then project image features onto the text space to guide the image encoder in acquiring ReID knowledge. Similar to CLIP-ReID \cite{li2023clip}, we utilize CLIP technology to align image and text features, thereby transferring the accumulated knowledge into the text space.

To complement the text modality, we further propose the Structured Semantic Prompt (SSP) learning and the Knowledge Adaptation and Projection (KAP) strategy to tackle the aforementioned challenges of knowledge mismatches in granularity and representation, respectively.
Concretely, we use learnable text prompts as containers for image-to-text distillation in the first stage, and decompose the prompt into several structured pairs. Each pair contains a set of tokens shared across all tasks for learning representation objects of person images, and a specific set of tokens for learning identity-related image representation content. With the explicit structured design, we can transfer knowledge between different tasks with uniform text description granularity. To alleviate the knowledge representation mismatch, we additionally tune the distilled text embeddings of different identities in the text semantic space during the second-stage learning. We optimize it with a slow-paced learner, so as to avoid knowledge confusion or forgetting and adjust for differences in representation between different tasks. 
Benefiting from the generalization of text representation and the proposed novel modules, our \textit{Teata} is rehearsal-free and capable of preserving old knowledge while adapting to new tasks for lifelong ReID of hybrid clothing conditions.

\noindent \textbf{Contributions.} We summarize our contributions as follows,

\textbf{(1)} We exploit a more practical lifelong ReID task, namely LReID with hybrid clothing states (LReID-Hybrid). It requires incrementally learning a ReID model under varying clothing states. To this end, we present a novel framework \textit{Teata}, which facilitates knowledge alignment, transfer, and accumulation in an ``image-text-image'' closed loop. 

\textbf{(2)} To address the limitation of knowledge granularity mismatch, we propose the Structured Semantic Prompt learning. It explicitly decomposes text prompts into a series of structured tokens, unifying the granularity of text representation and thereby enhancing knowledge transfer.

\textbf{(3)} We introduce a Knowledge Adaptation and Projection strategy to optimize the text representations of each identity using a slow-paced learner. This strategy not only adapts to the representation differences across tasks when guiding the learning of image encoders but also prevents knowledge forgetting concisely and effectively.

\textbf{(4)} Extensive experiments demonstrate the superiority of our proposed \textit{Teata} framework for LReID-Hybrid over various baselines. Additionally, it achieves state-of-the-art results with significant margins on conventional LReID benchmarks.

\section{Related Work}

\subsection{Person Re-Identification}
There has been remarkable progress in person re-identification leveraging pre-prepared stationary training data~\cite{wang2018learning,luo2019bag,zheng2019joint,wang2023rethinking}. Conventionally, researchers focus on the same-cloth scenario where people would always show a relatively stable clothing appearance. Based on this assumption, to reduce the influence of intra-class variations, Zheng \textit{et al.} \cite{zheng2019joint} propose to leverage the generated data and Zhou \textit{et al.} \cite{zhou2019omni} propose to perform omni-scale feature learning.
Wang \textit{et al.} \cite{wang2018learning} design a multi-branch deep network architecture integrating discriminative information with various granularities.
Some methods turn to auxiliary data clues~\cite{jin2020semantics,he2021transreid} for help. 
Wang \textit{et al.}~\cite{wang2023rethinking} rethink the role of the classifier as an interactive ReID process and propose a strong baseline from a projection-on-prototypes perspective.
Considering the huge annotation cost of identity labels, unsupervised person ReID has also received widespread attention~\cite{lin2020unsupervised,liu2022unsupervised,meng2024unleashing}. Following a clustering-based training strategy, Liu \textit{et al.} further propose a stochastic learning strategy and propose a unified distance matrix during clustering, and Lin \textit{et al.} generate style-transferred training images with different camera styles to achieve across-camera similarity exploration.

Although researchers have made efforts to promote the discriminative ability of person ReID models~\cite{wang2018learning,luo2019bag,zheng2019joint,wang2023rethinking,zhou2019omni,jin2020semantics,he2021transreid,lin2020unsupervised,liu2022unsupervised,meng2024unleashing}, these models would still fail when pedestrians change their clothes, since they almost exclusively focus on clothes.
Xu~\textit{et al.}~\cite{xu2021adversarial} and Eom \textit{et al.}~\cite{eom2021disentangled} attempt to use generative models to augment samples by explicitly synthesizing person images with various clothes, so as to learn more robust features against clothing changes.
Some studies draw support from auxiliary modalities to assist learning soft-biometrics features unrelated to clothes~\cite{qian2020long,yang2019person,wang2022co,jin2022cloth,hong2021fine,chen2021learning,liu2023dual,guo2023semantic,agbodike2023face,cui2023dcr}. For instance, SPT+ASE~\cite{yang2019person} utilizes reliable and discriminative curve patterns on the body contour sketch. 
CAMC~\cite{wang2022co} uses heatmaps of human postures to encode body shape semantic information. GI-ReID~\cite{jin2022cloth} learns cloth-agnostic representations by leveraging personal unique and cloth-independent gait information. 
Other researchers take advantage of clothing labels to eliminate the negative effects of clothing features. For instance, CAL~\cite{gu2022clothes} adopts adversarial learning to penalize the model's predictive power to clothes. AIM~\cite{yang2023good} adopts a dual-branch model to simulate causal intervention and eliminate clothing bias. 
Considering using well-trained off-the-shelf tools for modality extraction or manually annotated clothing labels can be inflexible, some studies \cite{wang2024exploring,wang2025content} focus on learning robust representations for cloth-changing person ReID by leveraging the mined fine-grained attributes~\cite{wang2024exploring} and abundant semantics~\cite{wang2025content} from images.

Recent research has successfully developed models trained on either cloth-changing or same-cloth datasets. These models are capable of recognizing different pedestrians in scenarios with a consistent clothing state. However, they struggle to perform well across domains with different clothing states, as knowledge acquired from cloth-changing and same-cloth domains exhibits significant differences in representation and granularity. For instance, ignoring cloth-related information would harm performance in same-cloth scenarios, where clothing can be one of the most discriminative characteristics. The relatively coarse-grained knowledge acquired from the same-cloth domain struggles to help distinguish pedestrians in cloth-changing scenarios.

\subsection{Lifelong Person Re-Identification}
Lifelong learning seeks to maintain stable performance on old tasks while adapting the models to gain new knowledge~\cite{li2017learning,wu2019large}. Due to the expansion of smart surveillance systems, person ReID models also have to continually accumulate knowledge information from old domains and generalize well on new domains~\cite{pu2021lifelong}. To meet the demand, AKA~\cite{pu2021lifelong} adopts knowledge distillation baselines~\cite{li2017learning} to preserve acquired knowledge and maintains a learnable knowledge graph to adaptively update previous knowledge. However, due to significant domain variations, it struggles to retain old knowledge without access to previous data.
GwFReID~\cite{wu2021generalising} formulates a comprehensive learning objective for maintaining coherence during progressive learning. To further promote LReID, ranking consistency distillation~\cite{pu2023memorizing}, distribution-aware prototypes~\cite{xu2024distribution}, and long short-term knowledge consolidation~\cite{xu2024lstkc} are proposed.
Recently, rehearsal-based methods have achieved state-of-the-art results by saving a few exemplars in each domain. For instance, PTKP~\cite{ge2022lifelong} proposes a pseudo task knowledge preservation framework to alleviate the domain gap in the last Batch Normalization layer. KRC~\cite{yu2023lifelong} introduces a dynamic memory model for bi-directional knowledge transfers and a knowledge consolidation scheme. 
However, it is impractical to save person images due to privacy issues, and the distribution discrepancy in the image space can aggravate knowledge forgetting~\cite{wang2025distribution}.

Additionally, as is opposite to reality, all previous works assume each pedestrian would always show a cloth-consistent appearance in the lifelong evolution process. They overlook the influence of clothing state variation and tend to fail when clothing information is not useful to distinguish different people.
Differently, we explore a more practical lifelong setting where both same-cloth domains and cloth-changing domains are considered. We show that the consistency, compactness, and generalization of text semantics can help integrate different knowledge across domains with various clothing states. Our proposed method does not rely on replayed exemplars and relieves the drawback of accumulating knowledge only in the image space.

\begin{figure*}
    \centering
    \includegraphics[width=0.93\linewidth]{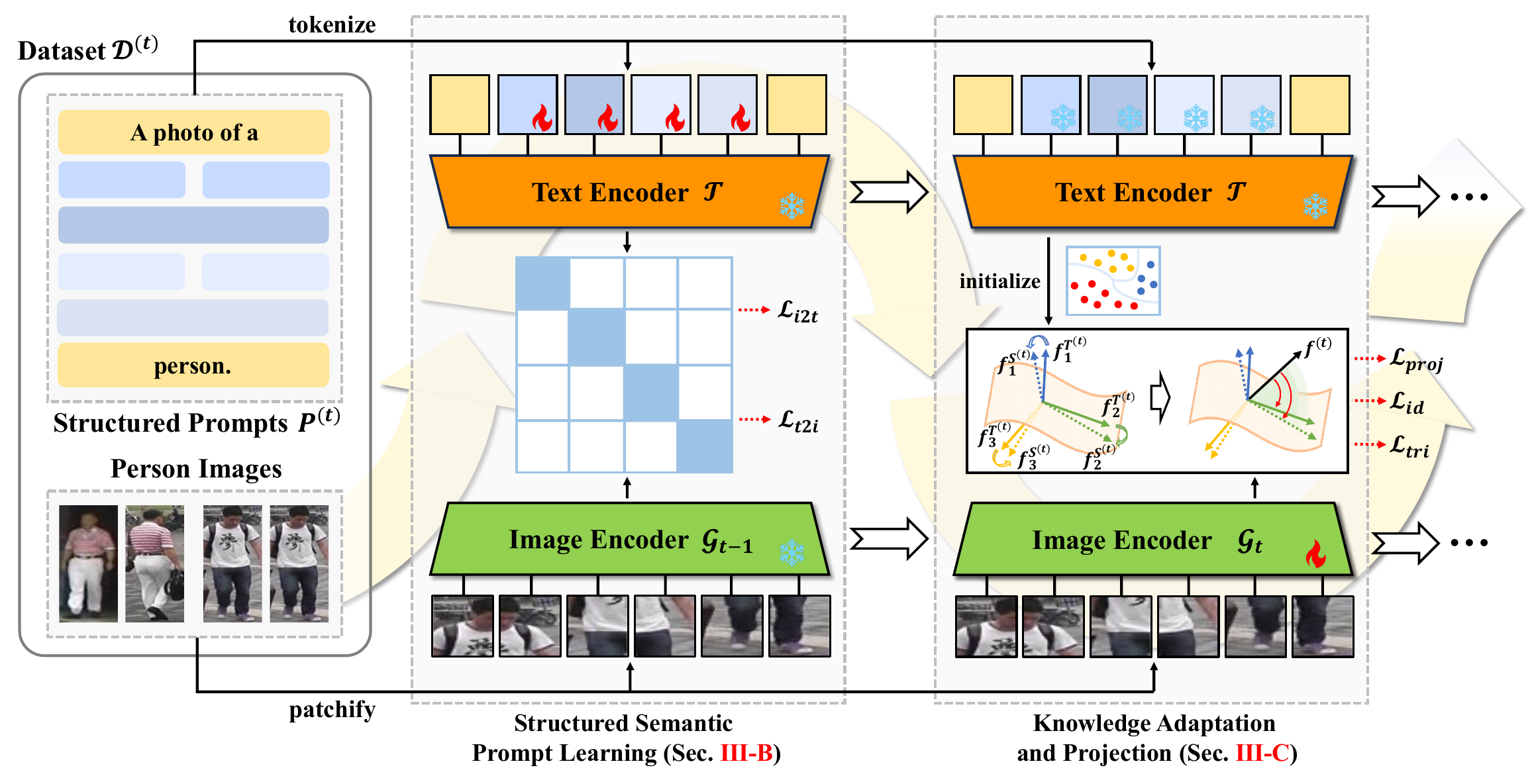}
    \caption{\textbf{Framework of our proposed \textit{Teata}.} 
    With the $t$-th training dataset $\mathcal{D}^{(t)}_{train}$, we first transfer visual information and category labels to the text space by aligning the image representations extracted by the image encoder $\mathcal{G}_{t-1}$ and the text embeddings extracted by the text encoder $\mathcal{T}$. 
    To achieve effective knowledge transfer, we design the Structured Semantic Prompt learning. It uses structured prompts to distill knowledge from the image space with a unified granularity of text description. Next, the distilled text embeddings are used to guide the evolution of the image encoder $\mathcal{G}_{t}$. For effective evolution, we introduce the Knowledge Adaptation and Projection strategy to tune text knowledge via a slow-paced learner and adapt $\mathcal{G}_{t}$ without catastrophic forgetting.}
    \label{fig:framework}
\end{figure*}

\section{Methodology}

\subsection{Formulation and Overview}
\subsubsection{Task Definition} 
In the task of lifelong person ReID with hybrid clothing states, a stream of datasets $\mathcal{D} = {\{ \mathcal{D}^{(t)} \}}^{T}_{t=1}$, where $T$ is the total number of datasets, are sequentially collected in different domains. $\mathcal{D}^{(t)}$ denotes the $t$-th added ReID dataset, which could be either \textit{the Same-Cloth scenario} (\textit{i.e.}, each pedestrian always wears the same clothes in different camera views), or \textit{the Cloth-Changing setting} (\textit{i.e.}, images of the same person can be captured appearing different clothing). 
It is a more practical yet challenging task than previous LReID \cite{ge2022lifelong,pu2021lifelong,yu2023lifelong}, since the model is required to perform well on seen or unseen domains with various clothing conditions, pursuing the ability to not only solve the catastrophic forgetting problem, but also balance the learned cloth-related and cloth-irrelevant identity knowledge.
At the $t$-th step, $\mathcal{D}^{(t)}=\{ \mathcal{D}^{(t)}_{train}, \mathcal{D}^{(t)}_{test}\}$ contains the training set and the testing set, each of which includes a variety of pairs of image $x_{i}^{(t)}$ and identity label $y_{i}^{(t)}$. An image encoder $\mathcal{G}_{t-1}$ from the $(t-1)$-th step is thus trained on the training set, to incrementally acquire new knowledge on the $t$-th dataset, that is, $\mathcal{G}_{t-1} \to \mathcal{G}_{t}$. Notably, samples of the previous datasets are not available at the current step due to data privacy and storage limitations. 
Eventually, the ReID model $\mathcal{G}_{T}$ should perform well on all seen domains, which is evaluated on the testing sets $\{\mathcal{D}^{(t)}_{test}\}^{T}_{t=1}$ at the end of the lifelong learning process of totally $T$ steps. Meanwhile, $\mathcal{G}_{T}$ should generalize well on the testing sets of unseen domains $\{\mathcal{D}^{(u)}_{u\_test}\}^{U}_{u=1}$, where $U$ is the total number of unseen datasets with various clothing states.

\subsubsection{Method Overview}
We propose a new framework \textit{Teata}, to effectively align, transfer, and accumulate knowledge using the text space as an intermediate medium. The overview of \textit{Teata} is illustrated in Fig.~\ref{fig:framework}. Inspired by CLIP-ReID \cite{li2023clip}, we adopt CLIP~\cite{radford2021learning} to bridge the image and text latent space, and build a closed loop of ``image-text-image''. We start by aligning image features with text embeddings to transfer visual information and category labels to the text space. Next, we leverage the robustness and consistency of text semantics to guide the discriminative feature learning of visual representations. This text-to-image knowledge transfer is finally reflected in the parameter update of the image encoder. 

Based on these two alternating stages, knowledge is continuously transferred in the image and text space, thereby achieving the accumulation of knowledge. To achieve effective knowledge transfer, we further design a Structured Semantic Prompt (SSP) learning in the first stage, which decomposes the original text prompt into several structured pairs to distill knowledge from the image space with a unified granularity of text description (see Sec.~\ref{subsec: prompt_learning}).
To further alleviate the knowledge confusion caused by incremental learning from same-cloth ReID and cloth-changing ReID tasks, we introduce a Knowledge Adaptation and Projection strategy (KAP) in the second stage, which tunes text knowledge via a slow-paced learner and adapts the image encoder to different tasks without catastrophic forgetting (see Sec.~\ref{subsec: classifier}). Lastly, we describe the details of training and testing procedures of \textit{Teata} in Sec.~\ref{subsec: procedure}.

\subsection{Structured Semantic Prompt Learning}
\label{subsec: prompt_learning}
CLIP-ReID \cite{li2023clip} utilizes a series of learnable tokens ${[X]}_{1}{[X]}_{2}\dots {[X]}_{M}$ as a container to distill visual knowledge from images. It is a free-from and task-driven learning process, as the features of each token learned could be different depending on different tasks or datasets. Although this token design is flexible, it could have a negative impact when performing lifelong learning across ReID datasets with different clothing states.
Knowledge acquired from same-cloth datasets tends to focus on the visual appearance features, while it requires more fine-grained discriminative clues, such as shoe style, gender, and body shape, when applying ReID models to the cloth-changing scenario. This mismatch, in terms of knowledge granularity, would cause a radically different update direction for each token, resulting in an inability to accumulate and transfer knowledge effectively.

Given a person image, we can describe its identity with text, for example, ``a middle age man with short hair wearing a white shirt and black pants''. We observe that the description of attributes has a similar form of [\textit{content object}]. Moreover, [middle age] and [short hair] are fine-grained identity information helpful for both same-cloth and cloth-changing ReID, while [white shirt] and [black pants] are cloth-related appearance information that is more discriminative under the condition of same-cloth ReID. It thus motivates us to present a Structured Semantic Prompt (SSP) with shared and specific tokens. Shared tokens are expected to function as [\textit{object}] and be shared across all tasks so that the text description has similar granularity. The specific tokens represent [\textit{content}] information unique to each identity, tailored for different ReID tasks. It is an ideal intuition, but unfortunately limited by the absence of attribute labels. To this end, we build upon learnable tokens and further structurally decompose them in the text semantics.
More specifically, we assign a structured text prompt $P_{j}^{(t)}$ to each pedestrian, where $j \in [1, N^{(t)}]$ and $N^{(t)}$ denotes the number of identities in the $t$-th dataset.
\begin{equation}
    P_{j}^{(t)} = {\rm A \ photo \ of \ a} \ [{X}_{1} \ {Y}_{1}] \cdots [{X}_{M} \ {Y}_{M}] \ {\rm person.} ,
\end{equation}
where $M$ means the total number of token pairs, ${X}_{m}$ and ${Y}_{m}$ $(m \in [1, M])$ denote the specific and shared tokens, respectively. They are all learnable parameters with the same dimension and are alternately combined to form each structured pair $[{X}_{m} \ {Y}_{m}]$. 
As discussed above, $\left\{X_{m}\right\}_{m=1}^{M}$ are not only dataset-specific but also identity-specific, that is, different people have different $\left\{X_{m}\right\}_{m=1}^{M}$. On the other hand, $\left\{Y_{m}\right\}_{m=1}^{M}$ are reused for all identities from different datasets.

With our proposed structured semantic prompt, models can integrate knowledge across domains with various clothing states in the unified textual semantic space. Analyses in Sec.~\ref{sec:ablation} and Tab.~\ref{tab:ablation} further support its efficacy, albeit in a straightforward form. To optimize it, we adopt CLIP contrastive losses $\mathcal{L}_{i2t}$ and $\mathcal{L}_{t2i}$~\cite{radford2021learning,li2023clip} which constrain the bidirectional alignment of image features and text embeddings:
\begin{equation}
    \mathcal{L}_{stage1} = \mathcal{L}_{i2t} + \mathcal{L}_{t2i}.
    \label{eq:astp}
\end{equation}

Overall, we use learnable text prompts as containers for image-to-text distillation, and decompose the prompt into several structured pairs. This is an ``implicitly feature learning with explicitly structural design'' approach to unify knowledge from different domains with the same granularity.

\subsection{Knowledge Adaptation and Projection}
\label{subsec: classifier}
Once the first stage of structured semantic prompt learning is done, we expect to transfer the knowledge from text to image space, so as to update the parameters of the image encoder $\mathcal{G}_{t}$. However, as shown in Fig.~\ref{fig:framework}, the previous model of $\mathcal{G}_{t-1}$ is used to extract features of samples from the $t$-th dataset, and then distill knowledge to text embeddings. In the LReID-Hybrid task, the knowledge acquired from ReID tasks with different clothing states may be contradictory. For example, the visual appearance features are discriminate representations for same-cloth ReID, but may be misleading for cloth-changing ReID. This inevitably leads to learned text embeddings being sub-optimal or not applicable, thus affecting effective knowledge transfer and accumulation. To alleviate the challenge of knowledge representation mismatch, we improve the second stage of learning with the Knowledge Adaptation and Projection (KAP) strategy.

\subsubsection{Adaptation with Slow-Paced Learner}
After the first stage, text embeddings for each identity can be obtained using the learned structured semantic prompts and the frozen text encoder. Formally, 
\begin{equation}
\mathbf{T}^{(t)} = \Bigg\{ {f^{T}_{j}}^{(t)} = \mathcal{T}\big(P_{j}^{(t)}\big) \ \bigg | \ j \in \left[1, N^{(t)} \right] \Bigg\},
\end{equation}
where $\mathcal{T}$ denotes the text encoder. As discussed above, instead of freezing text embeddings $\mathbf{T}^{(t)}$, we propose to fine-tune them with a slow-paced learner. It shows two benefits. First, we can continuously update text embeddings to make them more suitable for the current ReID task. Meanwhile, the slow-paced learner can prevent the learning of text embeddings from overfitting to the current task, resulting in forgetting previous knowledge. When updating text embeddings, the text encoder $\mathcal{T}$ is kept frozen and the structured semantic prompts are tuned for knowledge adaptation.

\subsubsection{Project-Then-Learn} 
To transfer the knowledge from text to image space effectively, we use the text embeddings to guide the learning of $\mathcal{G}_{t}$ by projecting image features on the text space and then updating the ReID model, \textit{i.e.}, the image encoder, to accumulate new knowledge. Formally,
\begin{equation}
    \mathcal{L}_{id} = - \sum\limits_{j = 1}^{N^{(t)}} q_{j} \ \log \frac{\exp \big({f^{(t)}_{i}}^{\top} \cdot {{f^{S}_{j}}^{(t)}}\big)}{\sum\limits_{k=1}^{N^{(t)}} \exp \big({f^{(t)}_{i}}^{\top} \cdot {{f^{S}_{k}}^{(t)}}\big)},
    \label{eq:v2t_id}
\end{equation}
\begin{equation}
    q_{j} = 
    \left\{
    \begin{aligned}
      &1 - \epsilon + \frac{\epsilon}{N^{(t)}} &&, \ j = y_{i}^{(t)} \\  
      &\frac{\epsilon}{N^{(t)}} &&, \ {\rm otherwise}
    \end{aligned}
    \right. ,
    \label{eq:q}
\end{equation}
where $f_{i}^{(t)}$ means the image feature of the $i$-th image with the identity label $y_{i}^{(t)}$, $i \in [1, n^{(t)}]$, and $n^{(t)}$ is the total number of the images in $\mathcal{D}^{(t)}_{train}$. ${f^{S}}^{(t)}$ is the adapted text embeddings that is initialized with the L2 normalized text embeddings $\mathbf{T}^{(t)}$, and ${f_{j}^{S}}^{(t)}$ denotes the adapted text embedding of the $j$-th identity. $\epsilon$ is a small constant for label smoothing regularization~\cite{inception_v3}, which is simply set to 0.1. 
Notably, Eq.~\ref{eq:v2t_id} is effective in fine-tuning the image encoder $\mathcal{G}_{t}$ together with text embeddings. As a result, with the help of the slow-paced learner, both image encoder and text embeddings can be slowly updated so that new knowledge is accumulated without forgetting the old. 

Since the $\mathcal{L}_{id}$ loss uses the adapted text embeddings, the learned text semantics from the first stage can be gradually lost. Therefore, we also use the $\mathcal{L}_{proj}$ loss~\cite{li2023clip}, which keeps the learned text embeddings frozen and uses them to guide the learning of the image encoder. $\mathcal{L}_{proj}$ can alleviate the loss of text semantics in the knowledge adaptation and projection process. It is formulated as:
\begin{equation}
    \mathcal{L}_{proj} = - \sum\limits_{j=1}^{N^{(t)}} q_{j} \ \log \frac{\exp \big({f^{(t)}_{i}}^{\top} \cdot {{f^{T}_{j}}^{(t)}}\big)}{\sum\limits_{k=1}^{N^{(t)}} \exp \big({f^{(t)}_{i}}^{\top} \cdot {{f^{T}_{k}}^{(t)}}\big)},
    \label{eq:proj}
\end{equation}
where ${f_{j}^{T}}^{(t)}$ is the text embedding of the $j$-th identity.

\subsubsection{Alternative Variant}
Benefiting from CLIP and our proposed SSP, text and image semantics are aligned in the shared latent space. Intuitively, image representations that are well aligned with text embeddings can also be used in Eq.~\ref{eq:v2t_id} to guide knowledge transfer and accumulation. Specifically, we can average all the extracted image features of each identity to obtain image representations $\mathbf{V}^{(t)}$ at the beginning of the second stage training. Formally,
\begin{equation}
   \mathbf{V}^{(t)} = {\left \{ {f^{V}_{j}}^{(t)} = \frac{1}{\left |X^{(t)}_{j} \right |} \sum\limits_{f^{(t)} \in X^{(t)}_{j}} f^{(t)} \ \Bigg | \ j \in \left [1, N^{(t)} \right ]  \right \}},
\end{equation}
where $X^{(t)}_{j}$ denotes the set of all person images with the same identity label $j$, and $|*|$ denotes the cardinal number of the set. Then, L2 normalized $\mathbf{V}^{(t)}$ is used for the initialization of the classifier in Eq.~\ref{eq:v2t_id}. In Sec.~\ref{sec:ablation}, we thoroughly investigate the efficacy of our designs, including knowledge adaptation with text embeddings (\textit{w/} KA-T), text-aligned image representations (\textit{w/} KA-I), random initialization (by default), and slow-paced learner (\textit{w/} SL).

\subsection{Details of Training and Inference}
\label{subsec: procedure}
Overall, two stages are involved in each step $t$ during the lifelong evolution of the model. In the first stage, both image encoder $\mathcal{G}_{t-1}$ and text encoder $\mathcal{T}$ are frozen, and Eq.~\ref{eq:astp} is adopted for the optimization of SSP. In the second stage, $\mathcal{G}_{t}$ is optimized with the guidance of the learned text embeddings. The losses involved in the second stage are formulated as:
\begin{equation}
    \mathcal{L}_{stage2} = \lambda_{1} \mathcal{L}_{proj} + \lambda_{2} \left(\mathcal{L}^{b}_{id} + \mathcal{L}_{id}\right) + \lambda_{3} \left(\mathcal{L}^{b}_{tri} + \mathcal{L}_{tri}\right),
\end{equation}
wherein $\mathcal{L}^{b}_{id}$ and $\mathcal{L}^{b}_{tri}$ denote the identity classification loss~\cite{luo2019bag} and triplet loss~\cite{hermans2017defense} applied on the image features before the last projecting layer of the CLIP image encoder. $\mathcal{L}_{id}$ and $\mathcal{L}_{tri}$ are applied to the representations after projection. $\lambda_{1}$, $\lambda_{2}$ and $\lambda_{3}$ are coefficients to balance between different losses. After training in step $t$, only the image encoder $\mathcal{G}_{t}$ is used for ReID evaluation. The cosine distances between two person images as measured during inference.

\section{Experiments}
\subsection{Datasets and Evaluation Protocol}

\subsubsection{Seen Datasets}
Two widely-used conventional same-cloth datasets, \textit{i.e.}, Market-1501~\cite{market1501} and MSMT17~\cite{wei2018person}, and two widely-used cloth-changing datasets, \textit{i.e.}, LTCC~\cite{qian2020long} and PRCC~\cite{yang2019person}, are used in our proposed LReID-Hybrid benchmarks for training. The dataset statistics are shown in Tab.~\ref{tab:dataset_statistics}. We design six training orders to imitate different situations comprehensively as follows,

\textit{Order 1:} Market-1501 $\rightarrow$ LTCC $\rightarrow$ MSMT17 $\rightarrow$ PRCC.

\textit{Order 2:} Market-1501 $\rightarrow$ LTCC $\rightarrow$ PRCC $\rightarrow$ MSMT17.

\textit{Order 3:} Market-1501 $\rightarrow$ MSMT17 $\rightarrow$ LTCC $\rightarrow$ PRCC.

\textit{Order 4:} LTCC $\rightarrow$ Market-1501 $\rightarrow$ PRCC $\rightarrow$ MSMT17.

\textit{Order 5:} LTCC $\rightarrow$ Market-1501 $\rightarrow$ MSMT17 $\rightarrow$ PRCC.

\textit{Order 6:} LTCC $\rightarrow$ PRCC $\rightarrow$ Market-1501  $\rightarrow$ MSMT17.

We also evaluate our proposed method on the widely-used standard benchmarks involving only datasets without clothing changes~\cite{wu2021generalising,ge2022lifelong}, where Market-1501~\cite{market1501}, DukeMTMC-reID~\cite{zheng2017unlabeled}, CUHK-SYSU~\cite{xiao2017joint} and MSMT17~\cite{wei2018person} are used sequentially for lifelong learning. DukeMTMC-reID is only used for academic use and fair comparison without identifying or showing pedestrian images. CUHK-SYSU is modified from the original for person search and rearranged following~\cite{wu2021generalising}. For other datasets, we follow the training and evaluation protocols proposed in their original papers. 

\subsubsection{Unseen Datasets} 
To evaluate the generalization of a model in the same-cloth scenario, we select five conventional ReID datasets, including CUHK01~\cite{li2013human}, CUHK02~\cite{li2013locally}, GRID~\cite{loy2010time}, SenseReID~\cite{zhao2017spindle} and PRID~\cite{hirzer2011person}. We follow their standard protocols to conduct the query and gallery sets for testing. As for the cloth-changing scenario, we evaluate lifelong ReID models on VC-Clothes~\cite{wan2020person} and Celeb-reID~\cite{huang2019beyond}, neither of which appeared in the training set. For the Celeb-reID dataset, we use its lightweight version (\textit{i.e.}, Celeb-reID-light) without losing generality, and each pedestrian in Celeb-reID-light does not wear the same clothes twice.

\begin{table}[t]
 \centering
  \caption{\textbf{The statistics of our LReID-Hybrid training benchmarks.} 
  }
  \label{tab:dataset_statistics}
  \setlength{\tabcolsep}{3.6mm}{
    \begin{tabular}{lrrrc}
    \toprule
    Dataset Names & \#IDs & \#images & \#cameras & SC \\
    \midrule
    Market-1501~\cite{market1501} & 1,501 & 32,668 & 6 & $\checkmark$ \\
    MSMT17~\cite{wei2018person} & 4,101 & 126,441 & 15 & $\checkmark$ \\
    LTCC~\cite{qian2020long} & 152 & 17,119 & 12 & $\scalebox{0.75}{\usym{2613}}$ \\
    PRCC~\cite{yang2019person} & 221 & 33,698 & 3 & $\scalebox{0.75}{\usym{2613}}$ \\
    \bottomrule
    \end{tabular}}
\end{table}

\subsubsection{Evaluation Metrics}
We adopt mean Average Precision (mAP) and Rank-1 accuracy (R-1) for performance evaluation on seen domains so far after each training step. We also report the average accuracies by averaging mAP and R-1 in seen domains as well as the accuracies in unseen domains. Considering the great performance discrepancy between same-cloth and cloth-changing datasets and to evaluate the model's ability in different clothing states, we report the average accuracies on seen same-cloth and cloth-changing datasets, respectively, which are denoted as $(\overline{s}_{\rm mAP}^{s}, \overline{s}_{\rm R-1}^{s})$ and $(\overline{s}_{\rm mAP}^{c}, \overline{s}_{\rm R-1}^{c})$. We denote the average accuracy on all unseen same-cloth and cloth-changing datasets as $(\overline{s}_{\rm mAP}^{us}, \overline{s}_{\rm R-1}^{us})$ and $(\overline{s}_{\rm mAP}^{uc}, \overline{s}_{\rm R-1}^{uc})$, respectively. For the LTCC, PRCC, and VC-Clothes datasets, we evaluate in the cloth-changing setting where only cloth-changing samples are involved in the testing set.

\subsection{Implementation Details}
We adopt the CLIP~\cite{radford2021learning} image encoder as the ReID model, which is ViT-B/16~\cite{dosovitskiy2020image} with 12 Transformer~\cite{vaswani2017attention} layers. The class token of the last layer is used as the visual representation of the input person image. The CLIP text encoder consists of 12 Transformer~\cite{vaswani2017attention} layers and the [EOS] token of the last layer is considered as the text embedding given structured prompts as input. 
The images are resized to 256 $\times$ 128 and augmented by random horizontal flipping, padding, cropping, and erasing~\cite{zhong2020random}. The batch size is set to 64, with 4 samples per pedestrian. For each domain, the model is optimized with Adam optimizer~\cite{kingma2014adam} with weight decay of $1 \times 10^{-4}$ in two stages. The first stage takes 120 epochs using a learning rate initialized as $3.5 \times 10^{-4}$ with cosine learning rate decay. The second stage takes 60 epochs with the warmup strategy that linearly increases the learning rate from $5 \times 10^{-7}$ to $5 \times 10^{-6}$ in the first 10 epochs. The learning rate is then decreased by a factor of 10 at the 40th epoch.
In our proposed KAP strategy, the slower-paced learner is adopted to reduce the learning rate by 10 times after the first domain. $M$ is set to 16. Following~\cite{li2023clip}, $\lambda_{1}$, $\lambda_{2}$, and $\lambda_{3}$ are set to 1, 0.25 and 1.

\begin{table*}[t]
  \centering
  \caption{\textbf{Comparison with the state-of-the-art methods in the LReID-Hybrid setting (Order 1).} The results are reported after the last training phase. The best results are shown in bold. }
  \label{tab:compare_order1}
    \setlength{\tabcolsep}{3.3mm}{
    \begin{tabular}{lcccccccccccc}
    \toprule
    {\multirow{2}[0]{*}{\textsc{Training Order 1}}}
    & \multicolumn{2}{c}{Market-1501} & \multicolumn{2}{c}{LTCC} & \multicolumn{2}{c}{MSMT17} & \multicolumn{2}{c}{PRCC} & \multicolumn{2}{c}{SC Average} & \multicolumn{2}{c}{CC Average}  \\
    \cmidrule(r){2-3} \cmidrule(r){4-5} \cmidrule(r){6-7} \cmidrule(r){8-9} \cmidrule(r){10-11} \cmidrule(r){12-13}
    & mAP & R-1 & mAP & R-1 & mAP & R-1 & mAP & R-1 & $\overline{s}_{\rm mAP}^{s}$ & $\overline{s}_{\rm R-1}^{s}$ & $\overline{s}_{\rm mAP}^{c}$ & $\overline{s}_{\rm R-1}^{c}$ \\
    \midrule
    AKA~\cite{pu2021lifelong} & 56.0 & 76.6 & 5.7 & 13.5 & 5.3 & 14.1 & 33.1 & 32.7 & 30.7 & 45.4 & 19.4 & 23.1 \\
    SFT & 54.8 & 76.3 & 16.0 & 34.2 & 45.5 & 72.0 & 47.4 & 47.0 & 50.2 & 74.2 & 31.7 & 40.6 \\
    LwF~\cite{li2017learning} & 62.8 & 81.4 & 17.1 & 31.9 & 53.3 & 77.6 & 47.5 & 46.3 & 58.1 & 79.5 & 32.3 & 39.1 \\
    CLIP-ReID~\cite{li2023clip} & 61.0 & 81.2 & 16.8 & 33.7 & 44.5 & 72.2 & 47.3 & 46.1 & 52.8 & 76.7 & 32.1 & 39.9 \\
    \midrule

    Teata (Ours) & \textbf{80.4} & \textbf{92.1} & \textbf{20.4} & \textbf{42.1} & \textbf{56.3} & \textbf{79.7} & \textbf{59.5} & \textbf{58.2} & \textbf{68.4} & \textbf{85.9} & \textbf{40.0} & \textbf{50.2} \\
    \midrule
    \rowcolor{gray!20} Joint-Train & 89.1 & 95.0 & 13.6 & 28.8 & 70.3 & 86.6 & 45.8 & 47.6 & 79.7 & 90.8 & 29.7 & 38.2 \\
    \bottomrule
    \end{tabular}}
\end{table*}

\begin{figure}[t]
    \centering
    \includegraphics[width=1\linewidth]{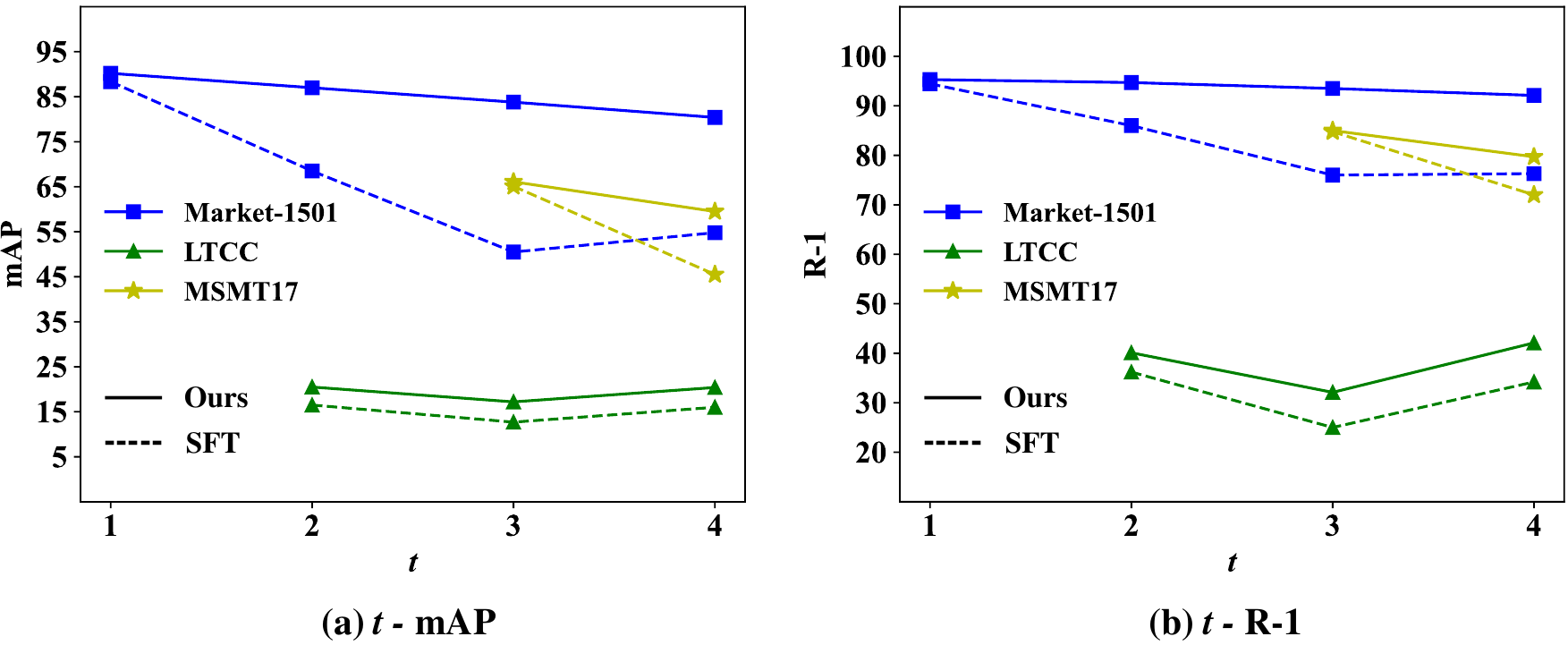}
    \caption{\textbf{Comparison results of our proposed method with SFT on each seen domain along the lifelong learning process.} We report the results in Order 1 w.r.t. (a) mAP and (b) R-1.}
    \label{fig:seen_acc}
\end{figure}

\begin{table*}[t]
  \centering
  \caption{\textbf{Comparison with the state-of-the-art methods in the LReID-Hybrid setting (Order 2)}. The results are reported after the last training phase. The best results are shown in bold. }
  \label{tab:compare_order2}
    \setlength{\tabcolsep}{3.3mm}{
    \begin{tabular}{lcccccccccccc}
    \toprule
    \multicolumn{1}{l}{\multirow{2}[0]{*}{\textsc{Training Order 2}}} 
    & \multicolumn{2}{c}{Market-1501} & \multicolumn{2}{c}{LTCC} & \multicolumn{2}{c}{PRCC} & \multicolumn{2}{c}{MSMT17} & \multicolumn{2}{c}{SC Average} & \multicolumn{2}{c}{CC Average}  \\
    \cmidrule(r){2-3} \cmidrule(r){4-5} \cmidrule(r){6-7} \cmidrule(r){8-9} \cmidrule(r){10-11} \cmidrule(r){12-13}
    & mAP & R-1 & mAP & R-1 & mAP & R-1 & mAP & R-1 & $\overline{s}_{\rm mAP}^{s}$ & $\overline{s}_{\rm R-1}^{s}$ & $\overline{s}_{\rm mAP}^{c}$ & $\overline{s}_{\rm R-1}^{c}$ \\
    \midrule
    AKA~\cite{pu2021lifelong} & 52.2 & 75.0 & 7.6 & 17.6 & 26.5 & 25.8 & 21.8 & 41.8 & 37.0 & 58.4 & 17.1 & 21.7 \\
    SFT & 49.5 & 74.5 & 11.6 & 25.3 & 22.3 & 20.8 & 63.8 & 84.7 & 56.7 & 79.6 & 22.8 & 23.1 \\
    LwF~\cite{li2017learning} & 58.7 & 79.6 & 14.2 & 27.6 & 32.1 & 32.0 & 60.2 & 82.1 & 59.5 & 80.9 & 23.2 & 29.8 \\
    CLIP-ReID~\cite{li2023clip} & 56.3 & 80.0 & 11.9 & 26.0 & 25.2 & 26.2 & \textbf{67.6} & \textbf{86.5} & 62.0 & 83.3 & 18.9 & 26.1 \\
    \midrule

    Teata (Ours) & \textbf{82.9} & \textbf{93.0} & \textbf{17.7} & \textbf{35.2} & \textbf{47.4} & \textbf{45.9} & 65.9 & 85.0 & \textbf{74.4} & \textbf{89.0} & \textbf{32.6} & \textbf{40.6} \\
    \midrule
    \rowcolor{gray!20} Joint-Train & 89.1 & 95.0 & 13.6 & 28.8 & 45.8 & 47.6 & 70.3 & 86.6 & 79.7 & 90.8 & 29.7 & 38.2 \\
    \bottomrule
    \end{tabular}}
\end{table*}

\begin{table*}[t]
  \centering
  \caption{\textbf{Comparison with the state-of-the-art methods in the LReID-Hybrid setting (Order 3).} The results are reported after the last training phase. The best results are shown in bold. }
  \label{tab:compare_order3}
    \setlength{\tabcolsep}{3.3mm}{
    \begin{tabular}{lcccccccccccc}
    \toprule
    {\multirow{2}[0]{*}{\textsc{Training Order 3}}}
    & \multicolumn{2}{c}{Market-1501} & \multicolumn{2}{c}{MSMT17} & \multicolumn{2}{c}{LTCC} & \multicolumn{2}{c}{PRCC} & \multicolumn{2}{c}{SC Average} & \multicolumn{2}{c}{CC Average}  \\
    \cmidrule(r){2-3} \cmidrule(r){4-5} \cmidrule(r){6-7} \cmidrule(r){8-9} \cmidrule(r){10-11} \cmidrule(r){12-13}
    & mAP & R-1 & mAP & R-1 & mAP & R-1 & mAP & R-1 & $\overline{s}_{\rm mAP}^{s}$ & $\overline{s}_{\rm R-1}^{s}$ & $\overline{s}_{\rm mAP}^{c}$ & $\overline{s}_{\rm R-1}^{c}$ \\
    \midrule
    AKA~\cite{pu2021lifelong} & 58.3 & 78.6 & 3.8 & 10.1 & 5.6 & 12.5 & 33.3 & 32.7 & 31.1 & 44.4 & 19.5 & 22.6 \\
    SFT & 54.8 & 77.0 & 37.0 & 65.7 & 16.6 & 35.5 & 46.9 & 48.2 & 45.9 & 71.4 & 31.8 & 41.9 \\
    LwF~\cite{li2017learning} & 63.9 & 83.0 & \textbf{49.3} & \textbf{75.3} & 17.8 & 31.9 & 46.9 & 44.6 & 56.6 & 79.2 & 32.4 & 38.3 \\
    CLIP-ReID~\cite{li2023clip} & 63.6 & 82.7 & 45.9 & 73.7 & 18.8 & 40.1 & 49.1 & 48.7 & 54.8 & 78.2 & 34.0 & 44.4 \\
    \midrule

    Teata (Ours) & \textbf{77.7} & \textbf{90.7} & 46.4 & 73.8 & \textbf{23.2} & \textbf{46.9} & \textbf{58.5} & \textbf{57.3} & \textbf{62.1} & \textbf{82.3} & \textbf{40.9} & \textbf{52.1} \\
    \midrule
    \rowcolor{gray!20} Joint-Train & 89.1 & 95.0 & 70.3 & 86.6 & 13.6 & 28.8 & 45.8 & 47.6 & 79.7 & 90.8 & 29.7 & 38.2 \\
    \bottomrule
    \end{tabular}}
\end{table*}

\begin{table*}[t]
  \centering
  \caption{\textbf{Comparison with the state-of-the-art methods in the LReID-Hybrid setting (Order 4).} The results are reported after the last training phase. The best results are shown in bold. }
  \label{tab:compare_order4}
    \setlength{\tabcolsep}{3.3mm}{
    \begin{tabular}{lcccccccccccc}
    \toprule
    {\multirow{2}[0]{*}{\textsc{Training Order 4}}} 
    & \multicolumn{2}{c}{LTCC} & \multicolumn{2}{c}{Market-1501} & \multicolumn{2}{c}{PRCC} & \multicolumn{2}{c}{MSMT17} & \multicolumn{2}{c}{SC Average} & \multicolumn{2}{c}{CC Average}  \\
    \cmidrule(r){2-3} \cmidrule(r){4-5} \cmidrule(r){6-7} \cmidrule(r){8-9} \cmidrule(r){10-11} \cmidrule(r){12-13}
    & mAP & R-1 & mAP & R-1 & mAP & R-1 & mAP & R-1 & $\overline{s}_{\rm mAP}^{s}$ & $\overline{s}_{\rm R-1}^{s}$ & $\overline{s}_{\rm mAP}^{c}$ & $\overline{s}_{\rm R-1}^{c}$ \\
    \midrule
    AKA~\cite{pu2021lifelong} & 11.4 & 25.0 & 27.9 & 51.3 & 29.6 & 32.6 & 11.5 & 27.1 & 19.7 & 39.2 & 20.5 & 28.8 \\
    SFT & 11.5 & 25.8 & 53.7 & 78.1 & 24.2 & 23.9 & 63.5 & {84.4} & 58.6 & 81.3 & 17.9 & 24.9 \\
    LwF~\cite{li2017learning} & 13.6 & 25.0 & 71.3 & 87.9 & 31.7 & 29.1 & 55.5 & 79.4 & 63.4 & 83.7 & 22.7 & 27.1 \\
    CLIP-ReID~\cite{li2023clip} & 12.0 & 24.0 & 61.1 & 82.5 & 23.0 & 23.3 & \textbf{66.2} & \textbf{86.0} & 63.7 & 84.3 & 17.5 & 23.7 \\
    \midrule

    Teata (Ours) & \textbf{17.4} & \textbf{37.2} & \textbf{72.4} & \textbf{88.9} & \textbf{48.6} & \textbf{49.6} & 64.9 & 83.9 & \textbf{68.7} & \textbf{86.4} & \textbf{33.0} & \textbf{43.4} \\
    \midrule
    \rowcolor{gray!20} Joint-Train & 13.6 & 28.8 & 89.1 & 95.0 & 45.8 & 47.6 & 70.3 & 86.6 & 79.7 & 90.8 & 29.7 & 38.2 \\
    \bottomrule
    \end{tabular}}
\end{table*}

\begin{table*}[t]
  \centering
  \caption{\textbf{Comparison with the state-of-the-art methods in the LReID-Hybrid setting (Order 5).} The results are reported after the last training phase. The best results are shown in bold. }
  \label{tab:compare_order5}
    \setlength{\tabcolsep}{3.3mm}{
    \begin{tabular}{lcccccccccccc}
    \toprule
    {\multirow{2}[0]{*}{\textsc{Training Order 5}}} 
    & \multicolumn{2}{c}{LTCC} & \multicolumn{2}{c}{Market-1501} & \multicolumn{2}{c}{MSMT17} & \multicolumn{2}{c}{PRCC} & \multicolumn{2}{c}{SC Average} & \multicolumn{2}{c}{CC Average}  \\
    \cmidrule(r){2-3} \cmidrule(r){4-5} \cmidrule(r){6-7} \cmidrule(r){8-9} \cmidrule(r){10-11} \cmidrule(r){12-13}
    & mAP & R-1 & mAP & R-1 & mAP & R-1 & mAP & R-1 & $\overline{s}_{\rm mAP}^{s}$ & $\overline{s}_{\rm R-1}^{s}$ & $\overline{s}_{\rm mAP}^{c}$ & $\overline{s}_{\rm R-1}^{c}$ \\
    \midrule
    AKA~\cite{pu2021lifelong} & 8.3 & 17.9 & 20.7 & 42.3 & 3.3 & 9.8 & 33.1 & 32.5 & 12.0 & 26.1 & 20.7 & 25.2 \\
    SFT & 14.4 & 29.3 & 61.2 & 81.4 & 44.9 & 72.1 & 52.0 & 54.7 & 53.1 & 76.8 & 33.2 & 42.0 \\
    LwF~\cite{li2017learning} & 14.6 & 28.6 & \textbf{75.8} & \textbf{89.2} & 48.2 & 74.3 & 44.4 & 41.9 & 62.0 & 81.8 & 29.6 & 35.3 \\
    CLIP-ReID~\cite{li2023clip} & 14.3 & 33.2 & 62.3 & 81.7 & 41.7 & 70.0 & 42.5 & 42.3 & 52.0 & 75.9 & 28.4 & 37.8 \\
    \midrule

    Teata (Ours) & \textbf{20.6} & \textbf{41.6} & 71.4 & 87.7 & \textbf{53.8} & \textbf{78.6} & \textbf{61.9} & \textbf{62.9} & \textbf{62.6} & \textbf{83.2} & \textbf{41.3} & \textbf{52.3} \\
    \midrule
    \rowcolor{gray!20} Joint-Train & 13.6 & 28.8 & 89.1 & 95.0 & 70.3 & 86.6 & 45.8 & 47.6 & 79.7 & 90.8 & 29.7 & 38.2 \\
    \bottomrule
    \end{tabular}}
\end{table*}

\begin{table*}[t]
  \centering
  \caption{\textbf{Comparison with the state-of-the-art methods in the LReID-Hybrid setting (Order 6).} The results are reported after the last training phase. The best results are shown in bold. }
  \label{tab:compare_order6}
    \setlength{\tabcolsep}{3.3mm}{
    \begin{tabular}{lcccccccccccc}
    \toprule
    {\multirow{2}[0]{*}{\textsc{Training Order 6}}} 
    & \multicolumn{2}{c}{LTCC} & \multicolumn{2}{c}{PRCC} & \multicolumn{2}{c}{Market-1501} & \multicolumn{2}{c}{MSMT17} & \multicolumn{2}{c}{SC Average} & \multicolumn{2}{c}{CC Average}  \\
    \cmidrule(r){2-3} \cmidrule(r){4-5} \cmidrule(r){6-7} \cmidrule(r){8-9} \cmidrule(r){10-11} \cmidrule(r){12-13}
    & mAP & R-1 & mAP & R-1 & mAP & R-1 & mAP & R-1 & $\overline{s}_{\rm mAP}^{s}$ & $\overline{s}_{\rm R-1}^{s}$ & $\overline{s}_{\rm mAP}^{c}$ & $\overline{s}_{\rm R-1}^{c}$ \\
    \midrule
    AKA~\cite{pu2021lifelong} & 11.0 & 22.4 & 28.5 & 30.0 & 27.6 & 54.3 & 10.3 & 25.5 & 19.0 & 39.9 & 19.8 & 26.2 \\
    SFT & 10.6 & 23.5 & 22.5 & 21.1 & 55.7 & 79.0 & 63.6 & 84.2 & 59.7 & 81.6 & 16.6 & 22.3 \\
    LwF~\cite{li2017learning} & 13.4 & 25.5 & 29.7 & 28.2 & 71.5 & 86.6 & 55.4 & 79.1 & 63.5 & 82.9 & 16.6 & 26.9 \\
    CLIP-ReID~\cite{li2023clip} & 11.7 & 23.5 & 26.2 & 25.4 & 61.9 & 83.2 & \textbf{65.7} & \textbf{85.3} & 63.8 & 84.3 & 19.0 & 24.5 \\
    \midrule

    Teata (Ours) & \textbf{16.5} & \textbf{36.0} & \textbf{45.4} & \textbf{44.6} & \textbf{75.1} & \textbf{90.0} & 64.9 & 84.2 & \textbf{70.0} & \textbf{87.1} & \textbf{31.0} & \textbf{40.3} \\
    \midrule
    \rowcolor{gray!20} Joint-Train & 13.6 & 28.8 & 45.8 & 47.6 & 89.1 & 95.0 & 70.3 & 86.6 & 79.7 & 90.8 & 29.7 & 38.2 \\
    \bottomrule
    \end{tabular}}
\end{table*}

\subsection{Comparison with the State-of-the-Art}
\noindent \textbf{Competitors.}
We first compare with a widely-used rehearsal-free method AKA~\cite{pu2021lifelong} for LReID. Then, we reimplement all competitors using CLIP~\cite{radford2021learning} pre-trained ViT-B/16~\cite{dosovitskiy2020image} as the ReID model.
Specifically, SFT denotes sequentially fine-tuning the model on each dataset, and Joint-Train denotes training on all datasets jointly, that is, all datasets are available to train the model simultaneously rather than sequentially. Joint-Train is traditionally regarded as the proximate performance upper bound of lifelong learning.
We also compare our proposed \textit{Teata} with LwF~\cite{li2017learning} and CLIP-ReID~\cite{li2023clip}. LwF performs knowledge distillation loss on the logit values of old tasks to keep task-specific decision boundaries. CLIP-ReID leverages the CLIP~\cite{radford2021learning} pre-trained text encoder and uses the learned text prompts to regularize the image encoder. CLIP-ReID is designed for conventional person ReID in a stationary domain, and we reimplement it to achieve lifelong learning by retraining text prompts on each sequentially coming dataset. 

\subsubsection{Results in Seen Domains}
We first show the results of all methods on our proposed LReID-Hybrid benchmarks. We comprehensively show the results of training with Orders 1$\sim$6 and evaluating on seen datasets in Tabs.~\ref{tab:compare_order1}$\sim$\ref{tab:compare_order6}. We also show results on the standard LReID benchmarks in Tab.~\ref{tab:compare_standard}.

\begin{table*}[t]
  \centering
  \caption{\textbf{Comparison with the state-of-the-art methods in the conventional LReID setting.} 
  ``\textit{w/} Ex.'' denotes rehearsal-based methods using exemplars. The training order is Market-1501 $\rightarrow$ DukeMTMC-reID $\rightarrow$ CUHK-SYSU $\rightarrow$ MSMT17, and the results are reported after the last training phase. The best results are shown in bold.}
  \label{tab:compare_standard}
    \setlength{\tabcolsep}{4mm}{
    \begin{tabular}{lccccccccccc}
    \toprule
    {\multirow{2}[0]{*}{Methods}} & \multirow{2}[0]{*}{\textit{w/} Ex.}
    & \multicolumn{2}{c}{Market-1501} & \multicolumn{2}{c}{DukeMTMC} & \multicolumn{2}{c}{CUHK-SYSU} & \multicolumn{2}{c}{MSMT17} & \multicolumn{2}{c}{SC Average}   \\
    \cmidrule(r){3-4} \cmidrule(r){5-6} \cmidrule(r){7-8} \cmidrule(r){9-10} \cmidrule(r){11-12}
    & & mAP & R-1 & mAP & R-1 & mAP & R-1 & mAP & R-1 & $\overline{s}_{\rm mAP}^{s}$ & $\overline{s}_{\rm R-1}^{s}$ \\
    \midrule
    
    AKA~\cite{pu2021lifelong} & & 59.7 & 80.1 & 32.7 & 48.3 & 82.0 & 84.4 & 17.1 & 34.9 & 47.9 & 61.9 \\
    GwFReID~\cite{wu2021generalising}  & $\checkmark$ & 60.9 & 81.6 & 46.7 & 66.5 & 81.4 & 83.9 & 25.9 & 52.4 & 53.7 & 71.1 \\
    LSTKC~\cite{xu2024lstkc} & & 56.9 & 78.0 & 54.3 & 70.5 & 84.4 & 86.0 & 34.0 & 60.3 & 57.4 & 73.7 \\
    DKP~\cite{xu2024distribution} & & 64.4 & 83.2 & 54.4 & 70.8 & 85.9 & 87.6 & 27.7 & 51.2 & 58.1 & 73.2 \\
    KRC~\cite{yu2023lifelong} & $\checkmark$ & 60.3 & 82.3 & 58.7 & 72.7 & 88.9 & 90.5 & 43.3 & 67.7 & 62.8 & 78.3 \\
    PTKP~\cite{ge2022lifelong} & $\checkmark$ & 75.8 & 89.7 & 62.0 & 76.7 & 85.0 & 86.3 & 34.5 & 60.9 & 64.3 & 78.4 \\
    \midrule
    LwF~\cite{li2017learning} & & 54.0 & 76.2 & 64.7 & 80.7 & 85.6 & 87.5 & 57.7 & 80.1 & 65.5 & 81.1 \\
    CLIP-ReID~\cite{li2023clip} & & 50.6 & 74.8 &60.0 & 76.8 & 84.5 & 86.4 & 64.7 & 84.9 & 65.0 & 80.7 \\
    \midrule
    
    Teata (Ours) & & \textbf{81.0} & \textbf{91.8} & \textbf{70.7} & \textbf{82.1} & \textbf{94.0} & \textbf{94.8} & \textbf{67.4} & \textbf{85.7} & \textbf{78.2} & \textbf{88.6} \\
    \midrule
    \rowcolor{gray!20} Joint-Train & & 89.8 & 95.1 & 81.2 & 90.6 & 95.1 & 95.8 & 72.3 & 87.3 & 84.6 & 92.2 \\
    \bottomrule
    \end{tabular}}
\end{table*}

\noindent \textbf{Results on LReID-Hybrid benchmarks (Order 1).} AKA~\cite{pu2021lifelong} achieves relatively poor results in the LReID-Hybrid task. Although AKA uses a knowledge graph to represent and accumulate knowledge, operating in the image space makes it difficult to maintain and accumulate knowledge when different tasks exhibit significant differences, especially with clothing state changes. Due to the absence of data from previous domains, SFT suffers greatly from catastrophic forgetting. LwF~\cite{li2017learning} performs knowledge distillation between current and frozen old models at the cost of sacrificing performance in the current domain. Due to the significant discrepancy across domains caused by variations in clothing states, LwF cannot effectively balance the knowledge from different domains.

Although CLIP-ReID~\cite{li2023clip} leverages text, it has no obvious advantage compared to LwF, which fine-tunes the pre-trained CLIP image encoder without the supervision of text. The results indicate that the potential of text is still not fully unleashed. In contrast, \textit{Teata} successfully leverages text semantics to balance and accumulate knowledge from various domains, demonstrating significant superiority over other methods. Note that \textit{Teata} does not require any task-specific tokens for inference, and the model parameters are shared across all tasks. Strictly speaking, in the setting of LReID-Hybrid, we do not have any clues about the next training task or the testing task, so task-specific tokens are not applicable. \textit{Teata} outperforms CLIP-ReID~\cite{li2023clip}, which learns text prompts as task-specific tokens at each training task, by a large margin.

It is also worth noting that the performance of Joint-Train is traditionally regarded as the proximate performance upper bound of the lifelong person ReID model. Similarly, when we train the standard and cloth-changing datasets jointly, we find the model can preserve its capability in the same-cloth scenario. However, the results in the cloth-changing domains are far from satisfactory. We argue that the acquired knowledge shows significant discrepancy due to various clothing states across domains. The data imbalance in distribution and quantity between standard and cloth-changing datasets causes the model to be unexpectedly biased toward the former when they are used jointly for training. The results further reveal the challenges faced in our proposed LReID-Hybrid task, which considers clothing variations during the real-world application of person ReID models. Unlike the image model, the text encoder is trained on a massive amount of data, building a compact and well-represented space. Our proposed \textit{Teata} leverages the text space as an intermediary for knowledge transfer and accumulation, effectively alleviating the representation bias caused by data imbalance.

In Fig.~\ref{fig:seen_acc}, we show the results in each seen domain along the lifelong learning process. Without the help of text semantics, SFT achieves inferior results in each domain. Worse still, when adapting to new domains, it forgets previously acquired knowledge dramatically. Thanks to the consistent and compact text semantics, \textit{Teata} can accumulate new knowledge while significantly reducing the loss of previously acquired knowledge. We also observe a slight performance improvement on the LTCC dataset when training on PRCC at step 4 compared with step 3. For \textit{Teata}, the final R-1 accuracy at step 4 is even higher than at step 2. We argue that since both PRCC and LTCC are cloth-changing datasets, the model tends to learn cloth-irrelevant identity representations on them. Therefore, the learned cloth-irrelevant identity clues on PRCC are useful in improving the discriminative ability for pedestrians with clothing changes on LTCC. \textit{Teata} exhibits a smoother forgetting trend in performance, showcasing its effectiveness in transferring and accumulating knowledge.

\noindent \textbf{More comparison results in other orders.} To verify the efficacy of our proposed framework \textit{Teata}, we perform comprehensive experiments in Orders 2$\sim$6 as shown in Tabs.~\ref{tab:compare_order2}$\sim$\ref{tab:compare_order6}. We compare one na\"ive solution of Sequential Fine-Tune (SFT), two rehearsal-free lifelong approaches AKA\cite{pu2021lifelong} and LwF~\cite{li2017learning}, as well as the CLIP-ReID~\cite{li2023clip} baseline which also uses text to promote ReID learning. Several observations can be made from the tables, demonstrating the effectiveness and robustness of our proposed method:
\textbf{(1)} \textit{Teata} consistently outperforms AKA, SFT, and LwF across different training orders by a large margin. This clearly highlights the advantage of leveraging the text space as an intermediary for knowledge transfer and accumulation across different domains.
\textbf{(2)} Although CLIP-ReID achieves relatively better results than other competitors in the same-cloth scenario, its performance degrades in the cloth-changing scenario. This may be due to i) the imbalance in the scale of same-cloth and cloth-changing datasets and ii) the mismatched knowledge representations for these two types of tasks, leading to catastrophic forgetting, especially in cloth-changing ReID. Our proposed designs of SSP learning and KAP strategy enable \textit{Teata} to successfully transfer and accumulate knowledge between same-cloth and cloth-changing domains, outperforming CLIP-ReID by a significant margin.
\textbf{(3)} Joint-Train generally performs better in the same-cloth scenario than in the cloth-changing scenario due to the discrepancy in different domains with various clothing states. In contrast, \textit{Teata} achieves a good balance in knowledge accumulation across domains, demonstrating its efficacy in lifelong learning.
\textbf{(4)} From the results of different training sequences, we observe that it is challenging to perform well in both cloth-changing and same-cloth domains simultaneously. Due to unavoidable knowledge forgetting, the model generally performs better in recently trained domains. Our proposed \textit{Teata} narrows the performance gap to some extent, showing about a 10\% improvement in $\overline{s}_{\rm mAP}$ and $\overline{s}_{\rm R-1}$ compared to SFT across all training orders and domains. 

\noindent \textbf{Results on conventional same-cloth LReID benchmarks.} Since our proposed method effectively handles the challenging LReID-Hybrid task, it is expected to also perform well when pedestrians do not change their clothes. In Tab.~\ref{tab:compare_standard}, we compare \textit{Teata} with state-of-the-art lifelong person ReID methods on the standard LReID benchmarks. Although LSTKC~\cite{xu2024lstkc} and DKP~\cite{xu2024distribution} adopt long short-term knowledge consolidation and distribution-aware prototypes, respectively, they still update the ReID models in the image space, facing the serious knowledge forgetting problem. Exemplars are considered vital in mitigating catastrophic forgetting, and replay-based methods~\cite{wu2021generalising,yu2023lifelong,ge2022lifelong} indeed perform better than AKA~\cite{pu2021lifelong}. However, saving person images can cause privacy issues and is not realistic in real-world applications. It also leads to significant resource consumption with the increase of training domains.
We also implement LwF~\cite{li2017learning} and CLIP-ReID~\cite{li2023clip} on the conventional LReID benchmarks. Leveraging the superior capability of CLIP, these methods achieve better results than state-of-the-art replay-based methods~\cite{wu2021generalising,yu2023lifelong,ge2022lifelong} on the current domain. However, their performance in previous domains is still unsatisfactory. In contrast, \textit{Teata} achieves a good balance across domains. With the help of 
consistent text semantics, we beat all competitors by a large margin without relying on impractical or inflexible exemplars.

It is worth noting that when comparing Order 2 of LReID-Hybrid with the conventional same-cloth LReID setting, we find little impact of cloth-changing datasets on same-cloth performance for \textit{Teata}. This demonstrates its effectiveness in handling knowledge mismatch and alleviating the knowledge forgetting caused by varying clothing states.

\begin{table*}[t]
\centering
\caption{\textbf{
Unseen domain comparisons with other competitors trained in the LReID-Hybrid setting (Order 1).} The results are given by the trained model after the last training step. ``$\dagger$'' means the light version of Celeb-reID. The best results are shown in bold.
}
\label{tab:unseen_order1}
\setlength{\tabcolsep}{1.35mm}{
    \begin{tabular}{lcccccccccccccccccc}
    \toprule
    {\multirow{2}[0]{*}{\textsc{Training Order 1}}} & \multicolumn{2}{c}{CUHK01} & \multicolumn{2}{c}{CUHK02} & \multicolumn{2}{c}{GRID} & \multicolumn{2}{c}{SenseReID} & \multicolumn{2}{c}{PRID} & \multicolumn{2}{c}{VC-Clothes} & \multicolumn{2}{c}{Celeb-reID$^\dagger$} & \multicolumn{2}{c}{SC Average} & \multicolumn{2}{c}{CC Average} \\
    \cmidrule(r){2-3} \cmidrule(r){4-5} \cmidrule(r){6-7} \cmidrule(r){8-9} \cmidrule(r){10-11} \cmidrule(r){12-13} \cmidrule(r){14-15} \cmidrule(r){16-17} \cmidrule(r){18-19}
    & mAP & R-1 & mAP & R-1 & mAP & R-1 & mAP & R-1 & mAP & R-1 & mAP & R-1 & mAP & R-1 & $\overline{s}_{\rm mAP}^{us}$ & $\overline{s}_{\rm R-1}^{us}$ & $\overline{s}_{\rm mAP}^{uc}$ & $\overline{s}_{\rm R-1}^{uc}$ \\
    \midrule
    SFT & 71.9 & 70.7 & 63.9 & 65.5 & 23.0 & 15.2 & 53.8 & 46.2 & 46.3 & 34.1 & 33.9 & 33.9 & 8.3 & 17.2 & 51.8 & 46.3 & 21.1 & 25.6 \\
    Joint-Train & 69.3 & 67.3 & 70.2 & 68.6 & 26.7 & 19.8 & 59.2 & 50.1 & 51.2 & 40.1 & 29.2 & 28.6 & 6.6 & 14.1 & 55.3 & 49.2 & 17.9 & 21.4 \\
    LwF~\cite{li2017learning} & 74.0 & 71.7 & 68.8 & 66.7 & 31.6 & 24.2 & 56.7 & 48.3 & 58.9 & 49.6 & 35.0 & 32.7 & 8.5 & 15.7 & 58.0 & 52.1 & 21.8 & 24.2 \\
    CLIP-ReID~\cite{li2023clip} & 76.3 & 75.2 & 71.6 & 72.4 & 31.7 & 24.3 & 56.6 & 48.0 & 54.2 & 43.9 & 39.4 & 37.5 & 9.0 & 17.2 & 58.1 & 52.8 & 24.2 & 27.4 \\
    \midrule
    
    Teata (Ours) & \textbf{84.0} & \textbf{82.6} & \textbf{77.0} & \textbf{75.3}  & \textbf{49.6}  & \textbf{41.6}  & \textbf{62.5}  & \textbf{54.2}  & \textbf{73.8}  & \textbf{65.0}  & \textbf{41.0}  & \textbf{40.2}  & \textbf{17.7}  & \textbf{35.3}  & \textbf{69.4} & \textbf{63.7} & \textbf{29.4} & \textbf{37.8} \\
    \bottomrule
    \end{tabular}}
\end{table*}

\begin{table*}[t]
\centering
\caption{\textbf{
Unseen domain comparisons with state-of-the-art models trained in the conventional LReID setting.} The training order is Market-1501 $\rightarrow$ DukeMTMC-reID $\rightarrow$ CUHK-SYSU $\rightarrow$ MSMT17. The results are given by the trained model after the last training step. ``$\dagger$'' indicates the light version of Celeb-reID. The best results are shown in bold.
}
\label{tab:unseen_standard}
\setlength{\tabcolsep}{1.55mm}{
    \begin{tabular}{lcccccccccccccccccc}
    \toprule
    {\multirow{2}[0]{*}{Methods}}  & \multicolumn{2}{c}{CUHK01} & \multicolumn{2}{c}{CUHK02} & \multicolumn{2}{c}{GRID}  & \multicolumn{2}{c}{SenseReID} & \multicolumn{2}{c}{PRID} & \multicolumn{2}{c}{VC-Clothes} & \multicolumn{2}{c}{Celeb-reID$^\dagger$} & \multicolumn{2}{c}{SC Average} & \multicolumn{2}{c}{CC Average} \\
    \cmidrule(r){2-3} \cmidrule(r){4-5} \cmidrule(r){6-7} \cmidrule(r){8-9} \cmidrule(r){10-11} \cmidrule(r){12-13} \cmidrule(r){14-15} \cmidrule(r){16-17} \cmidrule(r){18-19}
    & mAP & R-1 & mAP & R-1 & mAP & R-1 & mAP & R-1 & mAP & R-1 & mAP & R-1 & mAP & R-1 & $\overline{s}_{\rm mAP}^{us}$ & $\overline{s}_{\rm R-1}^{us}$ & $\overline{s}_{\rm mAP}^{uc}$ & $\overline{s}_{\rm R-1}^{uc}$ \\
    \midrule
    Joint-Train & 60.3 & 57.6 & 69.5 & 69.5 & 51.8 & 39.9 & 55.0 & 45.7 & 27.6 & 15.0 & 35.0 & 32.2 & 9.0 & 18.2 & 52.8 & 45.5 & 22.0 & 25.2 \\
    SFT & 75.6 & 74.7 & 61.6 & 58.4 & 32.0 & 21.9 & 52.9 & 44.4 & 49.0 & 35.8 & 33.0 & 31.4 & 6.0 & 13.6 & 54.2 & 47.0 & 19.5 & 22.5 \\
    PTKP~\cite{ge2022lifelong} & 71.1 & 69.4 & 66.4 & 66.1 & 33.9 & 25.3 & 53.9 & 46.1 & 38.4 & 27.3 & 24.2 & 21.8 & 5.3 & 9.2 & 52.7 & 46.8 & 14.8 & 15.5 \\
    KRC~\cite{yu2023lifelong} & 79.1 & 78.0 & 70.0 & 69.0 & 33.6 & 25.5 & 51.3 & 44.0 & 51.8 & 40.6 & 20.2 & 17.5 & 6.3 & 12.2 & 57.2 & 51.4 & 13.3 & 14.9 \\
    LwF~\cite{li2017learning} & 76.5 & 74.7 & 67.2 & 66.7 & 36.8 & 28.0 & 56.2 & 47.0 & 58.0 & 46.4 & 35.0 & 33.5 & 6.6 & 13.2 & 58.9 & 52.6 & 20.8 & 23.4 \\
    CLIP-ReID~\cite{li2023clip} & 77.2 & 75.6 & 65.3 & 64.9 & 35.4 & 25.3 & 56.7 & 48.5 & 53.3 & 40.5 & 36.4 & 34.9 & 6.0 & 11.4 & 57.6 & 51.0 & 21.2 & 23.2 \\
    \midrule
    
    Teata (Ours) & \textbf{86.9} & \textbf{84.6} & \textbf{82.2} & \textbf{81.4} & \textbf{54.9} & \textbf{45.6} & \textbf{67.5} & \textbf{58.3} & \textbf{80.7} & \textbf{72.0} & \textbf{40.5} & \textbf{39.2} & \textbf{11.3} & \textbf{24.1} & \textbf{74.4} & \textbf{68.4} & \textbf{25.9} & \textbf{31.7} \\
    \bottomrule
    \end{tabular}}
\end{table*}

\subsubsection{Results in Unseen Domains}
We also discuss another important capability of the lifelong ReID model, its generalization ability in unseen domains, given the continuous accumulation of knowledge across different ReID tasks. Without loss of generality, in Tabs.~\ref{tab:unseen_order1} and \ref{tab:unseen_standard}, we report results of models trained with Order 1 on the LReID-Hybrid task and on the standard LReID benchmarks in unseen domains, respectively. 

Similar to the conclusions from experiments in seen domains, Joint-Train performs better than SFT in the unseen same-cloth scenario but worse in the unseen cloth-changing scenario. The imbalance in the distribution and quantity of datasets biases the model towards clothing-related knowledge when different clothing-state datasets are used jointly for training. Additionally, LwF~\cite{li2017learning} and CLIP-ReID~\cite{li2023clip} improve capability in unseen domains through knowledge distillation and text modality. However, their performance remains unsatisfactory due to improper handling of knowledge representation and granularity mismatches.
Our method, as expected, outperforms all other competitors on all unseen datasets, demonstrating superior generalization ability. Thanks to our well-designed modules, \textit{Teata} aligns, transfers, and accumulates knowledge in an ``image-text-image" closed loop. Consistent and generalized text semantics effectively promote lifelong learning of ReID models, leading to superior results in both seen and unseen domains.

Furthermore, we train models only on the conventional LReID task and compare results on both same-cloth and cloth-changing unseen domains to explore the generalization ability of lifelong ReID models in greater depth. As shown in Tab.~\ref{tab:unseen_standard}, we observe the following:
\textbf{(1)} \textit{Teata} surpasses all competitors by a large margin in different scenarios. Despite not seeing any samples of changing clothes during training, it achieves impressive results in the cloth-changing scenario. This demonstrates the consistency and generalization of text semantics and the effectiveness of \textit{Teata}'s designs in utilizing text semantics and achieving knowledge accumulation in lifelong learning.
\textbf{(2)} Since only same-cloth datasets are used during training in the conventional LReID task, advanced ReID methods~\cite{ge2022lifelong,yu2023lifelong} achieve inferior results in unseen cloth-changing domains compared to those in Tabs.~\ref{tab:unseen_order1}.
\textbf{(3)} Advanced ReID methods~\cite{ge2022lifelong,yu2023lifelong} show a surprising decrease in performance in the cloth-changing scenario with improvement in the same-cloth scenario. These models may focus exclusively on clothing information due to overfitting on the training dataset bias, which is unreliable in the cloth-changing scenario.
\textbf{(4)} LwF~\cite{li2017learning} and CLIP-ReID~\cite{li2023clip} leverage the pre-trained vision-language model of CLIP. Benefiting from the complementarity and generalization advantages of image-text multimodality, these baselines perform better than previous lifelong ReID models~\cite{ge2022lifelong,yu2023lifelong}. However, their performance remains sub-optimal due to catastrophic forgetting caused by knowledge representation and granularity mismatches.
\textbf{(5)} Joint-Train shows no obvious advantage over other previous methods and may even perform worse on some same-cloth datasets like CUHK01~\cite{li2013human} and PRID~\cite{hirzer2011person}, while unexpectedly performing well in the unseen cloth-changing scenario. One possible reason is the mixture of data from different domains. Joint-Train tries to balance the domain gap, sacrificing the performance of datasets with large domain deviations or small scales. On the other hand, representation differences in clothing appearances across domains encourage the model to explore more robust representations beyond just clothing features. However, it still struggles with learning from different domains, resulting in limited performance.

\subsection{Ablation Studies}
\label{sec:ablation}

\begin{table}[t]
 \centering
  \caption{\textbf{Ablation studies of our proposed method.} The average accuracies of four datasets in Order 1 at the last training step are reported. 
  }
  \label{tab:ablation}
  \setlength{\tabcolsep}{2.5mm}{
    \begin{tabular}{lccccccc}
    \toprule
    \multirow{2}[0]{*}{No.} & \multirow{2}[0]{*}{SSP} & \multicolumn{2}{c}{KAP} & \multicolumn{2}{c}{SC Average} & \multicolumn{2}{c}{CC Average}  \\
    \cmidrule(r){3-4} \cmidrule(r){5-6} \cmidrule(r){7-8} 
    & & KA-T & SL & $\overline{s}_{\rm mAP}^{s}$ & $\overline{s}_{\rm R-1}^{s}$ & $\overline{s}_{\rm mAP}^{c}$ & $\overline{s}_{\rm R-1}^{c}$ \\
    \midrule
    1 & & & & 52.8 & 76.7 & 32.1 & 39.9 \\
    2 & $\checkmark$ & & & 58.3 & 80.7 & 33.8 & 42.8 \\
    3 & $\checkmark$ & $\checkmark$ & & 67.5 & \textbf{85.9} & 35.2 & 44.1 \\
    4 & $\checkmark$ & $\checkmark$ & $\checkmark$ & \textbf{68.4} & \textbf{85.9} & \textbf{40.0}  & \textbf{50.2} \\
    \bottomrule
    \end{tabular}}
\end{table}

\subsubsection{Effectiveness of SSP and KAP}
SSP transfers knowledge across the same-cloth and cloth-changing domains with a uniform granularity of text description. As shown in Tab.~\ref{tab:ablation}, it results in performance improvement in both same-cloth and cloth-changing scenarios. Furthermore, by addressing the knowledge representation mismatch problem through our KA-T and SL designs, our proposed \textit{Teata} achieves effective knowledge adaptation, leading to significant performance improvement, as evidenced in Tab.~\ref{tab:ablation}. It is worth mentioning that we also observe a similar trend in improving performance for each design in the traditional LReID setting without clothing state variations. SSP can leverage fine-grained attribute semantics in text descriptions by decomposing shared [object] information while keep the discriminative ability using decomposed [content] information. KAP prevents the learning of text embeddings from overfitting to the current task and thus avoids forgetting previous knowledge.

\begin{table}[t]
 \centering
  \caption{\textbf{Ablation results of knowledge adaptation in our KAP strategy.} The average accuracies of four datasets in Order 1 at the last training step are reported.
  }
  \label{tab:ablation_KR}
  \setlength{\tabcolsep}{4mm}{
    \begin{tabular}{lcccc}
    \toprule
    {\multirow{2}[0]{*}{Methods}} & \multicolumn{2}{c}{SC Average} & \multicolumn{2}{c}{CC Average}  \\
    \cmidrule(r){2-3} \cmidrule(r){4-5}
    & $\overline{s}_{\rm mAP}^{s}$ & $\overline{s}_{\rm R-1}^{s}$ & $\overline{s}_{\rm mAP}^{c}$ & $\overline{s}_{\rm R-1}^{c}$ \\
    \midrule
    Ours \textit{w/o} Align. & 63.0 & 82.5 & 34.5 & 43.9 \\
    Ours \textit{w/} KA-I & \textbf{69.0} & \textbf{86.1} & 39.5 & 49.2 \\
    Ours \textit{w/} KA-T & 68.4 & 85.9 & \textbf{40.0}  & \textbf{50.2} \\
    Ours \textit{w/o} $\mathcal{L}_{proj}$ & 64.4 & 83.5 & 37.7 & 46.5 \\
    \bottomrule
    \end{tabular}}
\end{table}

\subsubsection{Discussions of the KAP Strategy}
As discussed in Sec.~\ref{subsec: classifier}, since image and text modalities are aligned in the shared latent space, it would also be effective to use the image representations for knowledge adaptation. We verify this in Tab.~\ref{tab:ablation_KR} (``Ours \textit{w/} KA-I'') and observe similar results to ours. Furthermore, we remove the text alignment in the first stage and use the unaligned image representations for classifier initialization (``Ours \textit{w/o} Align.''). The significant performance decrease confirms the effectiveness of semantic alignment between image and text modalities. It suggests the advantage of the text space as an intermediary for knowledge transfer and accumulation across domains. We also ablate the $\mathcal{L}_{proj}$ loss in Tab.~\ref{tab:ablation_KR} (``Ours \textit{w/o} $\mathcal{L}_{proj}$''). The results demonstrate its effectiveness in strengthening the learning of the image encoder by alleviating the loss of text semantics in the knowledge adaptation and projection process.

\begin{table}[t]
 \centering
  \caption{\textbf{Comparison of one-stage and two-stage training strategies.} The average accuracies of four datasets in Order 1 at the last training step are reported.
  }
  \label{tab:ablation_stage}
  \setlength{\tabcolsep}{4mm}{
    \begin{tabular}{lcccc}
    \toprule
    {\multirow{2}[0]{*}{Methods}} & \multicolumn{2}{c}{SC Average} & \multicolumn{2}{c}{CC Average}  \\
    \cmidrule(r){2-3} \cmidrule(r){4-5}
    & $\overline{s}_{\rm mAP}^{s}$ & $\overline{s}_{\rm R-1}^{s}$ & $\overline{s}_{\rm mAP}^{c}$ & $\overline{s}_{\rm R-1}^{c}$ \\
    \midrule
    {One Stage} & {57.8} & {78.7} & {34.1} & {44.6} \\
    Two Stages (Ours) & \textbf{68.4} & \textbf{85.9} & \textbf{40.0}  & \textbf{50.2} \\
    \bottomrule
    \end{tabular}}
\end{table}

\subsubsection{Necessity of Two-Stage Training}
Our proposed two-stage iterative training takes advantage of semantics from the text encoder and shows great superiority in performance to the one-stage baseline SFT and the two-stage training baseline CLIP-ReID in Tabs. \ref{tab:compare_order1}$\sim$\ref{tab:unseen_standard}. Additionally, we also try a one-stage training variant, which adopts our proposed KAP strategy while uses the contrastive losses $\mathcal{L}_{i2t}$ and $\mathcal{L}_{t2i}$ to achieve our SSP learning at the same time. As shown in Tab. \ref{tab:ablation_stage}, the variant performs worse, since in the early stage of training the learned text prompts is ineffective in describing pedestrians. The text embeddings with misleading semantics would affect the KAP strategy to optimize the image encoder.

\begin{figure*}[t]
  \centering
  \includegraphics[width=0.85\linewidth]{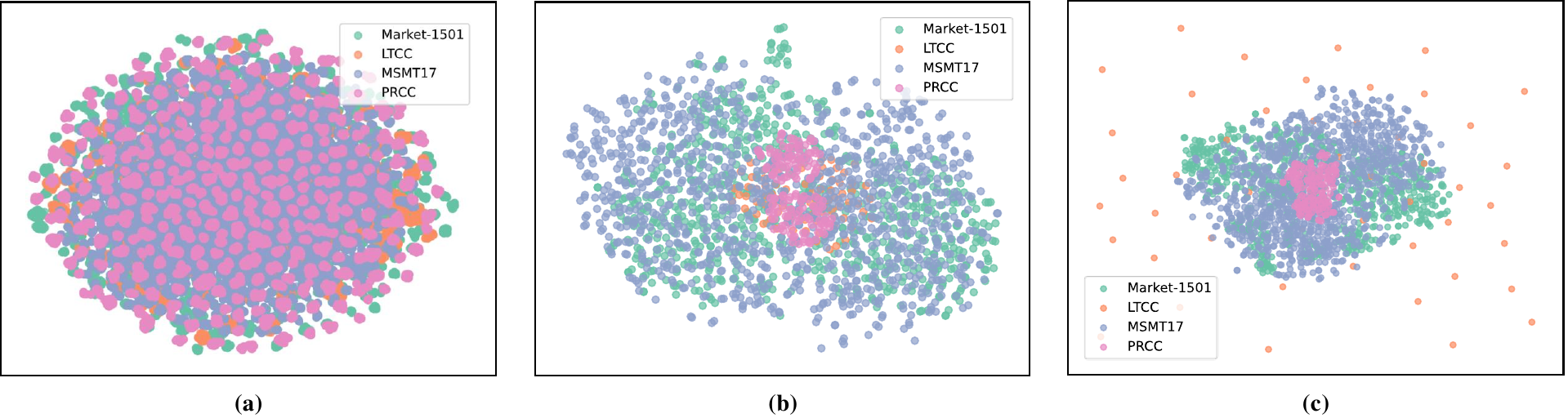}
  \caption{\textbf{Visualization of the learned representations in different domains with various clothing states.} For \textit{Teata}, we visualize the representations of (a) each image, and (b) each identity by averaging image representations with the same identity, respectively. We also visualize (c) the learned representations of each identity by SFT for comparison.}
  \label{fig:tsne}
\end{figure*}

\begin{figure}[tp]
  \centering
  \includegraphics[width=1\linewidth]{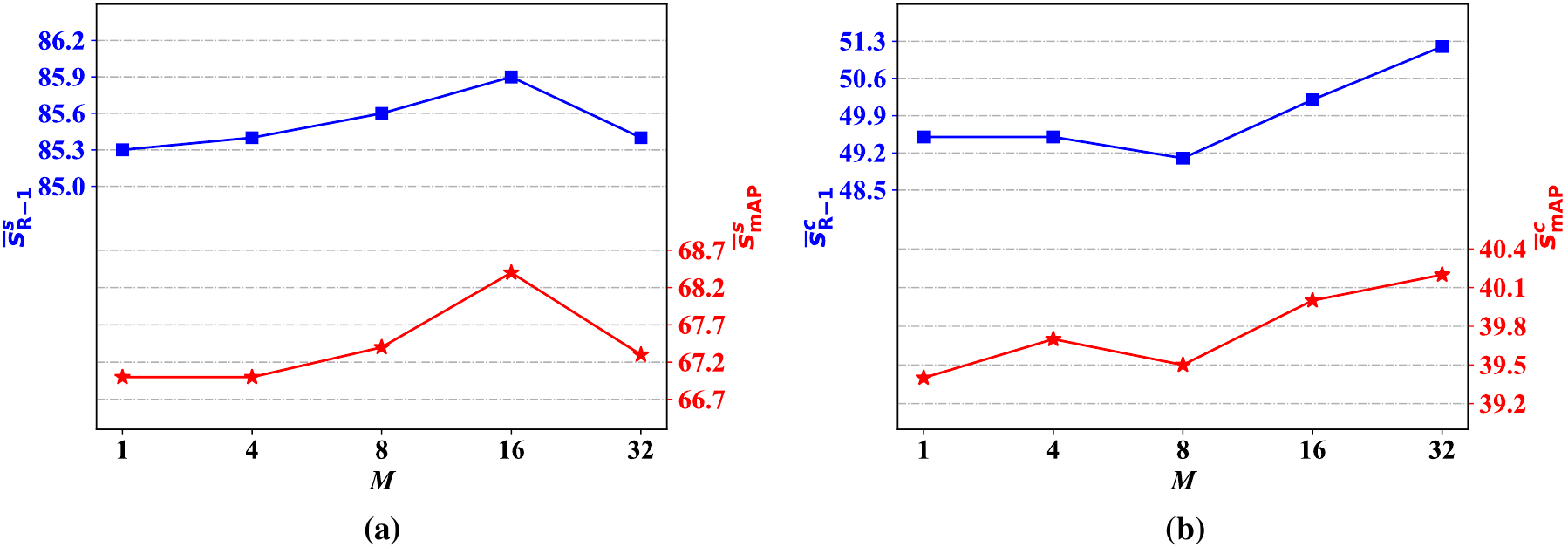}
  \caption{\textbf{Influence of different values of $M$}. The results are reported after the last training phase in Order 1.}
  \label{fig:M_ablation}
\end{figure}

\subsubsection{Influence of the Number of Token Pairs}
As shown in Fig.~\ref{fig:M_ablation}, a larger value of $M$ generally helps represent different attributes meticulously. We also find that a small number of token pairs can bring good performance. This can be attributed to the fact that the text space is more compact and generalized than the image space. The learnable structured tokens, acting as a container, can distill visual knowledge from images into higher-level representations. When we set $M$ to a large number, such as 32, we observe that it is beneficial in the cloth-changing scenario while providing less gain in the same-cloth scenario. This discrepancy arises from the knowledge granularity mismatch in different clothing states. As discussed in Sec.~\ref{sec:intro}, cloth-changing ReID typically requires models to perceive more fine-grained cues in person images, as visual appearance information, which dominates, can be misleading. To achieve a good balance between different knowledge requirements, we set $M=16$ for all our experiments.

\subsubsection{Visualizations of the Learned Representations}
Our proposed \textit{Teata} aligns image and text features, transferring the accumulated knowledge into the text space. Benefiting from the consistency, compactness, and generalization of text semantics, the learned representations are compact across domains. To demonstrate this, we use t-SNE~\cite{van2008visualizing} to visualize the learned representations in different domains. As shown in Fig.~\ref{fig:tsne} (a), the learned image representations are distributed in a unified latent space regardless of domains. The identity representations also exhibit great compactness in Fig.~\ref{fig:tsne} (b).
Upon closer observation of Fig.~\ref{fig:tsne} (b), the identity representations on the LTCC~\cite{qian2020long} and PRCC~\cite{yang2019person} datasets are distributed in a similar area. Since both involve clothing changes for each pedestrian, their identity representations show robustness to clothing variations, exhibiting cloth-irrelevant semantics. A similar phenomenon is observed on the Market-1501~\cite{market1501} and MSMT17~\cite{wei2018person} datasets, where each identity shows consistent appearances in clothing.
In contrast, as shown in Fig.~\ref{fig:tsne} (c), the learned identity representations of SFT show great discrepancy across different domains. The identity representations on Market-1501\cite{market1501}, MSMT17~\cite{wei2018person}, and PRCC~\cite{yang2019person} are distributed separately. Due to the great variations in clothing on LTCC~\cite{qian2020long}, the identity representations are dispersed. These inconsistent semantics can cause knowledge forgetting in the lifelong evolution process.
Overall, with the effective use of text semantics, our proposed \textit{Teata} enables the lifelong evolution of ReID models across domains and accumulates abundant knowledge to handle varying clothing states.

\subsection{More Discussions of the LReID-Hybrid Task}
We take a further step in LReID by  by taking mixed clothing states into account in LReID. For the cloth-changing domains, the newly added dataset may contain persons with different clothing states. For example, for the LTCC~\cite{qian2020long} training dataset, 46 people have clothing changes while 31 people always wear the same clothing. On the other hand, each person has more than one image, and they are randomly sampled during training. Therefore, there are both cloth-changing and same-cloth images when training in the cloth-changing domains.

To solve LReID-Hybrid task, we first propose SSP learning to transfer knowledge across domains with uniform text description granularity, and then introduce KAP strategy to guide the image representation learning via aligned text embeddings. It is worth noting that our method does not use any information or prior about the clothing states of the datasets.

\section{Conclusion}
Lifelong learning and handling clothing changes are two significant challenges in the application of person ReID. In this paper, we innovatively address both challenges simultaneously by proposing a more practical task: lifelong person re-identification with hybrid clothing states (LReID-Hybrid). This task considers a series of same-cloth and cloth-changing domains during the lifelong evolution of ReID models.
As pioneers in this area, we analyze the challenges posed by knowledge mismatches across domains with varying clothing states and explore the effectiveness of text for LReID-Hybrid. We leverage CLIP to take advantage of the robust alignment between image and text modalities. We further propose the Structured Semantic Prompt (SSP) learning and the Knowledge Adaptation and Projection (KAP) strategy to tackle the challenges of knowledge mismatches in granularity and representation, respectively. Our novel framework, \textit{Teata}, achieves knowledge alignment, transfer, and accumulation within an ``image-text-image" closed loop. Extensive experimental results on both LReID-Hybrid and conventional LReID benchmarks demonstrate the superiority of our approach.

We hope our work inspires more studies to focus on the lifelong evolution of ReID models with hybrid clothing states, thus advancing the deployment of ReID systems. Other issues, such as lifelong ReID across visible and infrared modalities, are also significant in real-world applications and remain underexplored. These scenarios present challenges comparable to LReID-Hybrid, and their solutions are far from trivial.
Additionally, our work has demonstrated the enhancing role of text in lifelong ReID, which hopes to encourage more research.

\bibliographystyle{IEEEtran}
\bibliography{egbib}

\end{document}